\PassOptionsToPackage{dvipsnames}{xcolor}

\documentclass{article} 
\usepackage[preprint]{colm2025_conference}

\usepackage{microtype}
\usepackage{hyperref}
\usepackage{url}
\usepackage{booktabs}

\usepackage{lineno}

\usepackage{graphicx}
\usepackage{tcolorbox}
\usepackage{amsmath}

\usepackage{bchart}
\usepackage{tikz}
 \usepackage{float}
\usepackage{amssymb}
\usepackage{colortbl}
\usepackage{pgfplots}
\usepackage{pgfplotstable}
\usepgfplotslibrary{groupplots} 
\usepackage[framemethod=TikZ]{mdframed}

\usepackage{tcolorbox}
\usepackage{enumitem}
\usepackage[utf8]{inputenc}
\usepackage[LFE,LAE]{fontenc}
\usepackage[english,farsi]{babel}

\usepackage{multirow}
\usepackage{pifont}
\usepgfplotslibrary{polar}
\pgfplotsset{compat=1.18}
\usetikzlibrary{positioning, shapes.geometric}

\pgfplotstableset{
	/color cells/min/.initial=0,
	/color cells/max/.initial=1000,
	/color cells/textcolor/.initial=,
	color cells/.code={%
		\pgfqkeys{/color cells}{#1}%
		\pgfkeysalso{%
			postproc cell content/.code={%
				\begingroup
				\pgfkeysgetvalue{/pgfplots/table/@preprocessed cell content}\value
				\ifx\value\empty
				\endgroup
				\else
				\pgfmathfloatparsenumber{\value}%
				\pgfmathfloattofixed{\pgfmathresult}%
				\let\value=\pgfmathresult
				\pgfplotscolormapaccess
				[\pgfkeysvalueof{/color cells/min}:\pgfkeysvalueof{/color cells/max}]
				{\value}
				{\pgfkeysvalueof{/pgfplots/colormap name}}%
				\pgfkeysgetvalue{/pgfplots/table/@cell content}\typesetvalue
				\pgfkeysgetvalue{/color cells/textcolor}\textcolorvalue
				\toks0=\expandafter{\typesetvalue}%
				\xdef\temp{%
					\noexpand\pgfkeysalso{%
						@cell content={%
							\noexpand\cellcolor[rgb]{\pgfmathresult}%
							\noexpand\definecolor{mapped color}{rgb}{\pgfmathresult}%
							\ifx\textcolorvalue\empty
							\else
							\noexpand\color{\textcolorvalue}%
							\fi
							\the\toks0 %
						}%
					}%
				}%
				\endgroup
				\temp
				\fi
			}%
		}%
	}
}

\definecolor{darkblue}{rgb}{0, 0, 0.5}
\hypersetup{colorlinks=true, citecolor=darkblue, linkcolor=darkblue, urlcolor=darkblue}

\title{MEENA (PersianMMMU): Multimodal-Multilingual Educational Exams for N-level Assessment}

\author{\small{Omid Ghahroodi$^{\triangle}$, Arshia Hemmat${}^{\spadesuit}$\thanks{These authors contributed equally to this work and are considered joint second authors. The order is listed randomly to reflect their equal contributions.} ,  Marzia Nouri${}^{\clubsuit*}$, Seyed Mohammad Hadi Hosseini${}^{\diamond*}$,}\\  \small{\textbf{Doratossadat Dastgheib${}^{\triangle*}$, Mohammad Vali Sanian${}^\diamond$\thanks{These authors contributed equally to this work and are considered joint third authors. The order is listed randomly to reflect their equal contributions.} , Alireza Sahebi${}^{\diamond\dagger}$, Reihaneh Zohrabi${}^{\diamond\dagger}$}}, \\  \small{\textbf{Mohammad Hossein Rohban${}^\diamond$\thanks{These authors contributed equally to this work and are considered joint corresponding authors. The order of corresponding authors is listed randomly to reflect their equal contributions.} , Ehsaneddin Asgari${}^{\triangle\ddagger}$, Mahdieh Soleymani Baghshah${}^{\diamond\ddagger}$}}
\\\\
$^{\diamond}$ Computer Engineering Department, Sharif University of Technology, Iran \\
$^{\triangle}$Qatar Computing Research Institute, Qatar\\
$^{\spadesuit}$ Computer Engineering Department, University of Isfahan, Iran\\
$^{\clubsuit}$ Independent Researcher
 \\\\
\small{\texttt{\{oghahroodi98, arshiahemmat6, nouri.marzia.1999, mvs2667\}@gmail.com}}\\
\small{\texttt{d\_dastgheib@sbu.ac.ir}, \texttt{easgari@hbku.edu.qa}}\\
\small{\texttt{\{hadi.hosseini17, alireza.sahebi, zohrabi, soleymani, rohban\}@sharif.edu}
}
}

\begin{document}
	
\selectlanguage{english}

\ifcolmsubmission
\linenumbers
\fi

\maketitle

\begin{abstract}
Recent advancements in large vision-language models (VLMs) have primarily focused on English, with limited attention given to other languages. To address this gap, we introduce MEENA (also known as PersianMMMU), the first dataset designed to evaluate Persian VLMs across scientific, reasoning, and human-level understanding tasks. Our dataset comprises approximately 7,500 Persian and 3,000 English questions, covering a wide range of topics such as reasoning, mathematics, physics, diagrams, charts, and Persian art and literature. Key features of MEENA include: (1) diverse subject coverage spanning various educational levels, from primary to upper secondary school, (2) rich metadata, including difficulty levels and descriptive answers, (3) original Persian data that preserves cultural nuances, (4) a bilingual structure to assess cross-linguistic performance, and (5) a series of diverse experiments assessing various capabilities, including overall performance, the model’s ability to attend to images, and its tendency to generate hallucinations. We hope this benchmark contributes to enhancing VLM capabilities beyond English.
\end{abstract}

\section{Introduction}
\label{sec:intro}
In recent years, vision-language models (VLMs) \citep{radford2021learningtransferablevisualmodels} have rapidly advanced, driving breakthroughs in multimodal tasks that integrate visual and textual understanding, such as visual question answering \citep{song2022clipmodelsfewshotlearners}, image captioning \citep{dai2023instructblipgeneralpurposevisionlanguagemodels}, embodied agents \citep{ma2025surveyvisionlanguageactionmodelsembodied} and document understanding \citep{luo2024layoutllm}. Despite their growing deployment, gaps in understanding VLMs' limitations highlight the need for comprehensive evaluation.

Several benchmarks have been developed to assess VLMs, each addressing different evaluation aspects. MMMU \citep{yue2024mmmu}, derived from college exams, quizzes, and textbooks, is designed to evaluate models on English-language exam questions. BLINK \citep{fu2024blinkmultimodallargelanguage} focuses on assessing models' performance on tasks that are intuitive for humans, such as visual similarity. MathVista \citep{lu2024mathvista} specializes in mathematical problem-solving and visual tasks, including tables and bar charts. AI2D \citep{kembhavi2016diagramworthdozenimages} facilitates question-answering based on diagrams, while MEGA-Bench \citep{chen2024megabenchscalingmultimodalevaluation} covers a diverse set of tasks, spanning coding, games, and scientific inquiries.
Despite progress in VLM evaluation, existing benchmarks remain predominantly English-centric. Moreover, linguistic and cultural differences underscore the necessity of benchmarks that are natively developed for each language rather than adapted through translation. This creates a pressing need for VLM benchmarks in languages beyond English, including Persian.

\begin{figure}[h]
    \centering
    \includegraphics[width=\textwidth]{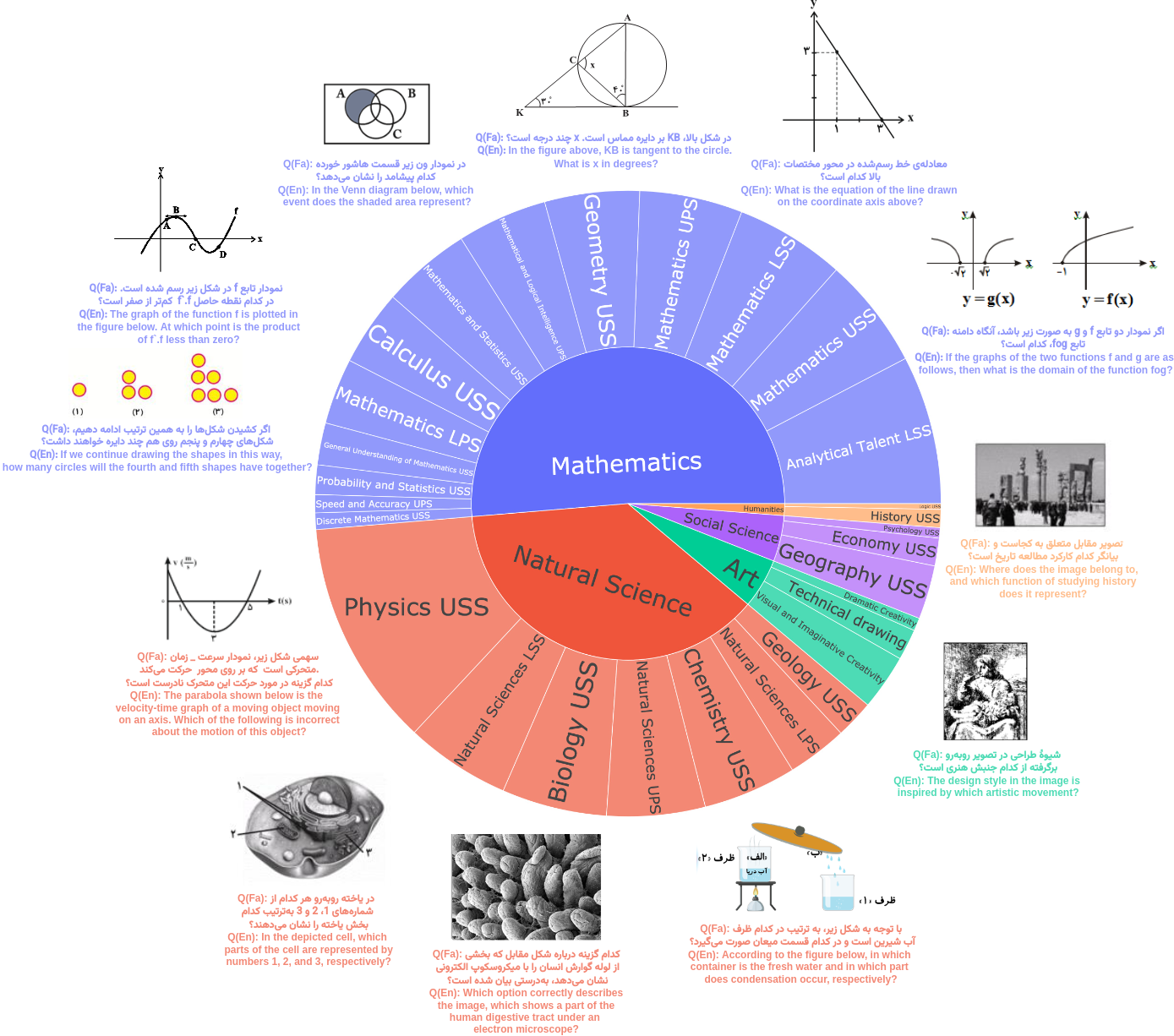}
    \caption{Overview of dataset and some sample questions from different tasks.}
\end{figure}

Persian benchmarks have largely concentrated on image captioning and visual question answering. However, they lack coverage of more complex human-level reasoning, such as mathematical problem-solving, spatial reasoning, and academic or scientific exam questions. Furthermore, most existing benchmarks focus primarily on text-based tasks and LLMs, rather than offering a comprehensive evaluation of VLMs.

Existing benchmarks for VLMs often prioritize limited aspects, such as perception or reasoning, mainly for high-resource languages like English, overlooking the challenges posed by diverse linguistic and cultural contexts.  In the case of Persian, a relatively low-resource language, these challenges are more pronounced, with available benchmarks failing to account for its unique linguistic features. \citet{farsi-etal-2025-persian} has several limitations, including its reliance on translations from English datasets, automated question generation, and a lack of complex, domain-specific reasoning tasks. Moreover, it does not evaluate reasoning in scientific disciplines such as mathematics or physics, which require structured logical thought. The use of translated datasets introduces cultural misalignment, which undermines the accuracy of evaluations. These limitations highlight the need for more comprehensive evaluation frameworks.

To address these gaps, we introduce \textbf{MEENA} (\textbf{M}ultimodal \textbf{E}ducational \textbf{E}xams for \textbf{N}-level \textbf{A}ssessment), the first dataset designed to evaluate Persian VLMs across scientific, reasoning, and human-level understanding tasks. The name MEENA (Mina) was chosen for this dataset due to its significance in Persian, where "Mina" refers to glass, and Mina-kari is a traditional art form. This aligns with the dataset’s multimodal nature, which includes a subset dedicated to art-related questions. Our dataset comprises approximately 7,500 Persian and 3,000 English questions, covering a wide range of topics such as reasoning, mathematics, physics, diagrams, charts, and Persian art and literature. This benchmark spans diverse subject areas across educational levels, from primary to upper secondary school, providing a comprehensive framework to assess the capabilities of VLMs. Key contributions of our work include:

\textbf{(1)} The first comprehensive Persian multimodal dataset for scientific and art exams, addressing the limitations of previous benchmarks by incorporating scientific and domain-specific tasks.

\textbf{(2)} Extensive experimentation covering a wide range of model evaluation scenarios, including Zero-Shot, Few-Shot, First Describe and focusing on the impact of visual input in different contexts (e.g., "Wronge Image" and "Without Image").

\textbf{(3)} A diverse set of questions varying in difficulty, topic, and format to test model across multiple contexts, such as reasoning, mathematics, physics, diagrams, charts, and Persian art and literature.

\textbf{(4)} Rich metadata, including difficulty levels, descriptive answers, and human performance enabling detailed performance analysis across various dimensions.

\textbf{(5)} Original Persian data that preserves cultural nuances, ensuring accurate evaluation in a culturally relevant context.

\textbf{(6)} A bilingual framework for cross-linguistic model evaluation, enhancing the assessment of performance in both Persian and English, and providing insights into cross-linguistic generalization.

Our dataset and code are available on HuggingFace and GitHub, respectively. Additionally, we set up a leaderboard on HuggingFace to stay updated with the performance of other models.

\section{Related Works}
\label{sec:related_works}
\subsection{Vision Language Models}

Multimodal vision-language models (VLMs) have emerged at the intersection of computer vision and natural language processing, allowing machines to interpret both visual and textual modalities \citep{li2025benchmark}. The limitations of large language models (LLMs) in handling single-modality data, particularly in capturing real-world information that requires multi-modal perception, have driven researchers to develop VLMs \citep{li2025surveystateartlarge, xu2024lvlm-ehub}. This has led to the rise of various models, including closed-source options like GPT-4o \citep{hurst2024gpt-4o}, Gemini \citep{team2023gemini}, and Claude \citep{claude2025}, as well as open-source models such as DeepSeek-VL2 \citep{wu2024deepseekVL2}, InstructBLIP \citep{dai2023instructblipgeneralpurposevisionlanguagemodels}, and Qwen2.5-VL \citep{bai2025qwen2.5VL}.
VLMs are increasingly applied in generative AI systems \citep{abootorabi2025generativeaicharacteranimation}, retrieval-augmented generation (RAG) systems \citep{abootorabi-etal-2025-ask}, education, and healthcare \citep{10.3389/frai.2024.1430984}.

\subsection{VLM Evaluation Benchmarks}


Despite significant advancements, current VLMs still struggle with certain categories of visual tasks, such as visual arithmetic \citep{huang2025why-vision} (including geometric problem-solving \citep{gao2023g-llava}) and spatial reasoning, which encompasses spatial relations, orientation, and navigation \citep{stogiannidis2025mindgap, chen2024spatialvlm}. To evaluate VLM performance in these areas, various benchmarks have been introduced. MMMU \citep{yue2024mmmu} provides a benchmark featuring multiple-choice and open-ended questions designed to assess VLMs’ perception, knowledge, and reasoning abilities. MMT-Bench \citep{ying2024mmt} evaluates VLMs across 32 tasks requiring expert knowledge, visual reasoning, and localization. Additionally, \citet{stogiannidis2025mindgap, chen2024spatialvlm} present benchmarks specifically focused on assessing VLMs’ spatial reasoning capabilities.


Despite existing benchmarks for VLM evaluation, few are designed to assess performance in low-resource languages such as Persian, particularly in scientific knowledge and visual reasoning. COCO-Flickr Farsi \citep{Coco-Flickr_Farsi} and Persian OCR dataset \citep{Persian_OCR_dataset} address image captioning and optical character recognition. ParsVQA-Caps \citep{mobasher2022parsvqa} provides a visual question answering task in which questions are generated using templates and by human annotators from images gathered from the web.

\citet{ghahroodi2024khayyam} introduces a large-scale, culturally grounded benchmark with 20,805 questions across 38 tasks, enabling rigorous and contamination-free evaluation of LLMs in Persian. Although it covers scientific and reasoning aspects, it lacks visual components and corresponding analysis. \citet{farsi-etal-2025-persian} provides a valuable benchmark with five distinct sets of questions, including visual abstract reasoning, word-image puzzles to assess models' ability to combine visual information with linguistic interpretation, and Iran-places to measures the models' knowledge of notable places in Iran. Despite these valuable questions, the benchmark lacks coverage of most aspects of scientific reasoning \citep{ma2024sciagent}. While it features abstract reasoning questions, it does not evaluate reasoning in many scientific domains such as mathematics or physics, which requires structured logical thinking. This limitation reduces its effectiveness in assessing models’ ability to solve complex scientific problems. Furthermore, the dataset lacks diversity in its task range, limiting the subjects on which models can be tested. Using non-original questions translated from English, and questions generated by LLMs to assess VLMs could lead to inaccurate evaluation of models in Persian language \citep{al2024analysisEducationalQuestions}. Additionally, since the dataset is closed-source, its applicability in evaluating other VLMs remains limited. A comparison of a number of Persian and English datasets is provided in Table \ref{compare_benchs}.

\begin{table*}
  \centering
  \scalebox{0.57}{
  \begin{tabular}{ccccccccc}
    \hline
     &  &  &  &  & \multicolumn{4}{c}{\textbf{Metadata}} \\
    
    \multirow{-2}{*}{\textbf{Dataset}} & \multirow{-2}{*}{\textbf{Languages}} & \multirow{-2}{*}{\textbf{VU Tasks}} & \multirow{-2}{*}{\textbf{Type \& \# Sample (Img)}} & \multirow{-2}{*}{\textbf{Access}}  & Desc. Ans. & Diff. Lev. & Trap & \# Tasks (Subtasks) \\
    \hline
    MMT-Bench & Eng. & MCQA & gen: 31.3K & Open & \ding{53} & \ding{53} & \ding{53} & 32 (162) \\
    MMMU & Eng. & MCQA, AM & orig: 11.5K & Open & 18\% & \ding{51} & \ding{53} & 30 (183) \\
    Farsi et. al. & Per., Eng. & MCQA, AM & tran: 7.7K(1K), gen: 70K(7K), orig: 0.6K+? & Closed & \ding{53} & \ding{53} & \ding{53} & 5 \\
    ParsVQA-Caps & Per. & AM, IC & orig: 27.5K(18.5K) & Open & \ding{53} & \ding{53} & - & 11 \\
    PICD* & Per. & IC & orig: 41K(1.5K) & Open & - & - & - & 5 \\
    CFF** & Per. & OCR & tran: 124K & Open & - & - & - & 1 \\
    Persian-OCR & Per. & OCR & orig: 33K & Open (7K) & - & - & - & 2 \\
    MEENA (Ours) & Per., Eng. & MCQA & orig(Per.): 7.4K (tran(Eng.): 3K) & Open & \ding{51} & \ding{51} & \ding{51} & 27 \\
    \hline
  \end{tabular}}
  \caption{\label{compare_benchs} A summary of various English and Persian VLM benchmarks, detailing the supported languages, vision understanding tasks, type and number of questions, accessibility, metadata (descriptive answers, difficulty levels, traps), and number of tasks (subtasks). As trapped questions are not defined for questions other than multiple choice, we marked those fields with a hyphen (-). The same is applied for descriptive answers of questions other than multiple choice and answer matching. VU: Vision Understanding, Desc. Ans.: Descripive Answer, Diff. Lev.: Difficulty Level, \# Img: Number of images, Eng: English, MCQA: Multiple Choice Question Answering, gen: Generated Questions, orig: Original Questions, tran: Translated Questions, IC: Image Captioning, OCR: Optical Character Recognition, AM: Answer Matching. In Answer Matching, the answer of question could be short-form, long-form, number, or yes/no.   \\ *Persian Image Captioning Dataset \citep{PICD}\\ **COCO-Flickr Farsi \citep{Coco-Flickr_Farsi}}
\end{table*}

\section{Dataset}
\label{sec:dataset}
The MEENA benchmark offers a robust collection of data designed to evaluate vision-language models (VLMs) with Persian language support, focusing on multiple-choice question answering. This dataset spans a wide array of disciplines and educational levels, assessing a range of cognitive skills, including reasoning, knowledge application, and comprehension. It is derived from Iran's 12-year educational framework, which consists of 6 years of primary education—divided into lower primary (LP, years 1–3) and upper primary (UP, years 4–6)—and 6 years of secondary education, split into lower secondary (LS, years 7–9) and upper secondary (US, years 10–12).

\subsection{Data Compilation}
The dataset primarily originates from two sources: (1) the ``Pellekan Yadgiri'' (Learning Ladder) platform, operated by the Kanoon Farhangi Amoozesh (Cultural Educational Institute) in Iran, which provides educational resources and standardized exercises, and (2) a curated selection of questions from online sources, including items from the Iranian national university entrance exams. The compilation process involved several steps:
\textbf{(1) Extraction and Cleaning}: Parsed HTML data to extract question attributes, removed questions with tables or explanatory answers, and deduplicated entries.
\textbf{(2) Image Processing}: Retained only questions with visual elements, categorized as: (i) questions with a single image, (ii) choices with a single image, or (iii) both question and choices with images. For cases with multiple images, these were merged into a single image to ensure compatibility across VLMs, as some models cannot process multiple inputs. Examples are provided in the appendix.
\textbf{(3) Content Filtering}: Excluded categories with insufficient visual questions (e.g., literature).
\textbf{(4) Diversity and Contribution}: Questions stem from a broad pool of educators, reducing individual bias and enhancing variety.
The dataset is licensed under Creative Commons No Derivatives (CC ND). Sampling was weighted using the formula \(1 / \text{weight}^{1/4}\), where \(\text{weight}\) denotes the number of questions per category, tuned to \(1/4\) to address data imbalance while preserving diversity. Uniform sampling or fixed-size subsets were avoided to maintain fairness across categories with varying question counts. A bilingual subset of 3,067 questions (547 from online sources and 2,520 from Pellekan Yadgiri) was created by translating items with Persian text in images into English, retaining only those with pure English or non-text visuals. Detailed examples of the MEENA dataset are included in Appendix \ref{app:dataset}.

\subsection{Metadata Details}
The Pellekan Yadgiri subset, forming the bulk of the MEENA benchmark, includes rich metadata to support in-depth analysis:
\begin{itemize}
    \item \textbf{Educational Level}: Tags questions to LP, UP, LS, or US, aligning difficulty with expected knowledge at each stage.
    \item \textbf{Difficulty Rating}: Assigns one of five levels—easy, moderately easy, medium, moderately hard, hard—for granular performance assessment.
    \item \textbf{Answer Explanations}: Provides detailed reasoning for each correct answer, aiding comprehension and evaluation.
    \item \textbf{Trap Indicators}: Flags questions with misleading ``trap'' choices, often rated as harder, to study reasoning pitfalls.
    \item \textbf{Student Success Rate}: Records the percentage of students answering correctly, offering a human performance baseline.
    \item \textbf{Subject Breakdown}: Organizes questions into precise topics (e.g., ``Mathematics → Algebra → Equations''), enabling targeted analysis.
    \item \textbf{Creation Year}: Tracks the year of question design, revealing trends in complexity over time.
\end{itemize}
This metadata enables comparisons between human and VLM performance, highlighting strengths and weaknesses in reasoning, trap avoidance, and topic-specific proficiency.

\subsection{Statistical Overview}
The dataset comprises 7,483 multiple-choice questions: 6,936 from Pellekan Yadgiri and 547 from online sources. It covers domains such as humanities, mathematics, sciences, and reasoning skills, with 6,936 questions linked to human performance data (Pellekan Yadgiri only) and a subset featuring trap elements. 

\subsection{Translation Process}
\label{sec:translation_process}

To extend our primarily Persian dataset into an English counterpart, we adopted a systematic translation pipeline that combines both automated methods and quality checks. Our main translation engine is \textbf{GPT-4o}, configured to handle multi-sentence and domain-specific text. 

\textbf{Evaluation Methodology: }
We applied an \textbf{LLM-as-a-Judge} approach, inspired by recent studies \citet{feng2024mmad,gu2025surveyllmasajudge,zhu2025judgelmfinetunedlargelanguage, zheng2023judgingllmasajudgemtbenchchatbot}, in which a large language model (GPT-4o in an evaluator mode) directly compares the translated text to the original Persian input. This model provides a semantic alignment score on a scale from 1 to 5, thus going beyond token matching to incorporate context-aware judgments about meaning preservation and fluency.

\textbf{Selection Criterion: }
All translated samples scoring \textbf{4 or higher} on the 1–5 scale were retained for the final English dataset. Samples below this threshold underwent additional review or revision to address discrepancies. This filtering ensures that only high-quality English renditions of Persian questions persist, resulting in a consistent, reliable dataset suitable for cross-lingual vision-language model evaluations.

\subsection{Distinguishing Features}
The MEENA benchmark excels due to:
\begin{itemize}
    \item \textbf{Broad Scope}: Encompasses diverse fields from analytical reasoning to scientific inquiry across educational stages.
    \item \textbf{Enhanced Metadata}: Offers contextual depth for sophisticated model evaluation.
    \item \textbf{Persian Authenticity}: Retains original Persian content with cultural relevance, avoiding translation artifacts.
\end{itemize}

\section{Experiments}
\label{sec:experiments}

\subsection{Experiment Overview}

We analyze two languages (Persian and English) and classify each question into three different cases based on the presence of images. Specifically:
\begin{itemize}
    \item \textbf{Questions with images:} Only the question prompt contains images.
    \item \textbf{Choices with an image:} Only the answer options (choices) contain images.
    \item \textbf{Both inquiries and selections involving pictures:} Images are present in both the question and its multiple-choice options.
\end{itemize}

We evaluate \textbf{GPT-4o} and \textbf{GPT-4o-mini} \citep{openai2024gpt4o}, \textbf{GPT-4-Turbo} \citep{openai2023gpt4turbo}, \textbf{Gemini-2.0-flash} \citep{google2023gemini}, and \textbf{InstructBLIP-T5} \citep{dai2023instructblip} on \textbf{Persian} and \textbf{English} data.

To determine how visual information affects the model's performance, each of these three cases is examined independently. We further design five experimental settings (Zero-Shot, In-Context Learning, First Describe, Wrong Image, Without Image) to isolate the impact of different multimodal cues and prompting strategies. for further details about the models and the rationality behind the experiment settings, see Appendix~\ref{app:experiment}.

\subsection{Experimental Design}
\label{sec:experimental_design}  

Below, we formalize each of our five main experiment types using a uniform notation. Let 
\(q_{*}\) be the textual question (in Persian or English), \(x_{*}\) be the (true or substituted) image relevant to \(q_{*}\), \(c_{*}\) be the correct answer or label we aim to predict, \(M(\cdot)\) denote the model’s output given specified inputs.
For every experiment, the same set of questions is used. 

\paragraph{Zero-Shot (ZS).}
A minimal-guidance setup in which the model receives only the single question-image pair \(\bigl(q_{*}, x_{*}\bigr)\) with no supplemental examples:
\[
   \hat{c}_{*} \;=\; M\bigl(q_{*},\, x_{*}\bigr).
\]
Here, \(\hat{c}_{*}\) represents the model’s direct output under default settings. Concretely, each input prompt includes the text of \(q_{*}\) and the raw image as two distinct inputs. No additional context (such as sample Q\&A pairs) is provided. Each pair is processed independently, ensuring no cross-contamination of information between different items.

\paragraph{In-Context Learning (ICL).}
We provide \(k\) example triplets \(\{(q_{i}, c_{i})\}_{i=1}^k\) as demonstrations immediately before the target query \((q_{*}, x_{*})\):
\[
   \hat{c}_{*} \;=\; M\Bigl(\bigl\{(q_{i}, c_{i})\bigr\}_{i=1}^k,\; q_{*}, x_{*}\Bigr).
\]
The value of \(k\) set to four and is kept consistent within each run. Additionally, the examples were chosen manually, ensuring the examples are informative and relevant to the questions topic.

\paragraph{First Describe (FD).}
\label{ss:first_describe}
We draw inspiration from chain-of-thought prompting approaches \citep{wei2023chainofthoughtpromptingelicitsreasoning} that encourage models to generate intermediate reasoning steps in text form before producing a final output. Similar works on multimodal reasoning \citep{
rose2024visualchainthoughtbridging,zhang2024multimodalchainofthoughtreasoninglanguage,zheng2024thinkinglookingimprovingmultimodal} also motivate explicit step-by-step analysis of visual content. In our adaptation, we create a form of ''visual chain of thought'' for each image, aiming to prevent the model from taking shortcuts (i.e., guessing an answer without fully accounting for the image).


\paragraph{Experiments with Mismatched or Missing Images.}
Before introducing the \emph{Wrong Image} and \emph{Without Image} settings, we note that prior research on multimodal grounding and visual-text alignment has explored techniques such as image substitution or omission to diagnose model dependencies \citep{hemmat2024hidden, favero2024multimodalhallucinationcontrolvisual, villa2025eagleenhancedvisualgrounding, gunjal2024detectingpreventinghallucinationslarge, wang2024mitigatinghallucinationslargevisionlanguage}. In our design, we follow similar practices to investigate whether the absence or irrelevance of the image affects a model’s predictive outcome. We adopt two settings that vary the presence or correctness of the accompanying image:

\paragraph{Wrong Image (WI).}
\label{ss:wrong_image}
We replace the correct image \(x_{*}\) with an intentionally mismatched or irrelevant image \(\hat{x}\) that does not correspond to \(q_{*}\):
\[
   \hat{c}_{*} \;=\; M\bigl(q_{*},\, \hat{x}\bigr).
\]
All other prompt elements remain unchanged. Each wrong image \(\hat{x}\) is drawn from a pool of images that are confirmed to be unrelated to the content of \(q_{*}\). This ensures the mismatch is unambiguous. The input format (text+image) is kept identical to Zero-Shot, except we swap out the image.

\paragraph{Without Image (WO).}
We remove the image entirely:
\[
   \hat{c}_{*} \;=\; M\bigl(q_{*},\, x_{*} = \varnothing\bigr).
\]
In practice, the model is given only the text of \(q_{*}\), and references to an image are either omitted or replaced with a placeholder (e.g., “[No Image Provided]”) depending on how the prompts are typically structured. The rest of the setup, including question style and domain, remains identical to Zero-Shot.

\subsection{Answer Extraction}
To assess model performance, it is essential to identify the option selected by the model in its generated response and use it to compute accuracy. To achieve this, we implement a two-stage framework. In the first stage, we apply regex-based pattern matching to extract explicit statements, such as "The correct answer is option 2." When these predefined rule-based patterns match, we can confidently extract the model’s selected option. However, in approximately half of the cases, regex patterns do not yield a match. Furthermore, depending on the nature of the experiment such as scenarios where no image is provided, the model may correctly infer the absence of an image and generate a response like "An image is required to answer this question." To handle such cases, we leverage LLM as a judge, utilizing the GPT-4o-mini model to infer the selected option when explicit patterns are absent. This model also determines whether the response indicates a missing image, assesses instances where the model fails to comprehend the question, and identifies responses that indicate an incorrect image reference.

All prompts used in the experiments, translations, and LLM as a judge are provided in the appendix \ref{app:prompts}.

\section{Results and discussions} 
\label{sec:results}
\begin{figure*}[h]
    \centering
    \resizebox{\textwidth}{!}{
    \begin{tikzpicture}
        \begin{groupplot}[
            group style={
                group size=2 by 1, 
                horizontal sep=3cm, 
            },
            width=10cm,
            height=8cm
        ]

        \nextgroupplot[
            xlabel={(a)},
            ylabel={Accuracy (\%)},
            xtick={1, 2, 3, 4},
            xticklabels={gpt-4o, gpt-4o-mini, gpt-4-turbo, gemini-2.0-flash},
            ymin=0.15, ymax=0.55,
            legend pos=north west,
            legend style={
                at={(0.0,1)},
                fill=none,
                legend columns=2,
                text width=3cm,
                draw=none,
                font=\small},
            ymajorgrids=true,
            grid style=dashed
        ]
        
        \addplot[color=blue!50,mark=triangle*,thick] coordinates {
            (1, 0.2996) (2, 0.2173) (3, 0.1901) (4, 0.3154)
        };
        \addlegendentry{Reasoning (fa)}
        
        \addplot[color=Emerald,mark=triangle*,thick] coordinates {
            (1, 0.4433) (2, 0.3097) (3, 0.3391) (4, 0.3962)
        };
        \addlegendentry{Knowledge (fa)}
        
        \addplot[color=blue!50,dashed,mark=*,thick] coordinates {
            (1, 0.3367) (2, 0.1810) (3, 0.3532) (4, 0.1773)
        };
        \addlegendentry{Reasoning (en)}
        
        \addplot[color=Emerald,dashed,mark=*,thick] coordinates {
            (1, 0.4859) (2, 0.3674) (3, 0.4599) (4, 0.3962)
        };
        \addlegendentry{Knowledge (en)}

        \node[anchor=north east] at (rel axis cs: 0.02, 1) {(a)};

        \nextgroupplot[
            xlabel={(b)},
            ybar,
            bar width=0.3cm,
            symbolic x coords={gpt-4-turbo, gpt-4o-mini, gpt-4o, gemini-2.0-flash},
            xtick=data,
            ymin=0, ymax=10,
            ylabel={\% of 'no\_image' Answers},
            ymajorgrids=true,
            x tick label style={rotate=0, anchor=center, yshift=-3pt},
            legend style={at={(0.5,1.0)},anchor=north,legend columns=-1},
            nodes near coords,
            every node near coord/.append style={font=\small},
        ]
        
        \addplot[ybar, draw=Emerald, fill=Emerald!60] coordinates {
            (gpt-4-turbo, 0.81) (gpt-4o-mini, 2.33) (gpt-4o, 1.76) (gemini-2.0-flash, 5.49)
        };
        
        \addplot[ybar, draw=blue, fill=blue!20, xshift=0.3cm] coordinates {
            (gpt-4-turbo, 1.49) (gpt-4o-mini, 2.72) (gpt-4o, 0.57) (gemini-2.0-flash, 9.17)
        };

        \legend{en, fa}

        \node[anchor=north east] at (rel axis cs: 0.02, 1) {(b)};

        \end{groupplot}

        \draw[dashed] (rel axis cs:1.25,0) -- (rel axis cs:1.25,1);

    \end{tikzpicture}
    }

    \caption{(a) Accuracy comparison of reasoning and knowledge-based tasks across models in English (en) and Persian (fa). Solid lines represent Persian tasks, while dashed lines indicate English tasks. (b) Comparison of 'no image' error rates for English (en) and Persian (fa) inputs. GPT-4-Turbo and GPT-4o maintain consistently low error rates in both languages, while Gemini 2.0 Flash exhibits significantly higher errors, particularly for Persian inputs.}
    \label{fig:grouped_plots}
\end{figure*}
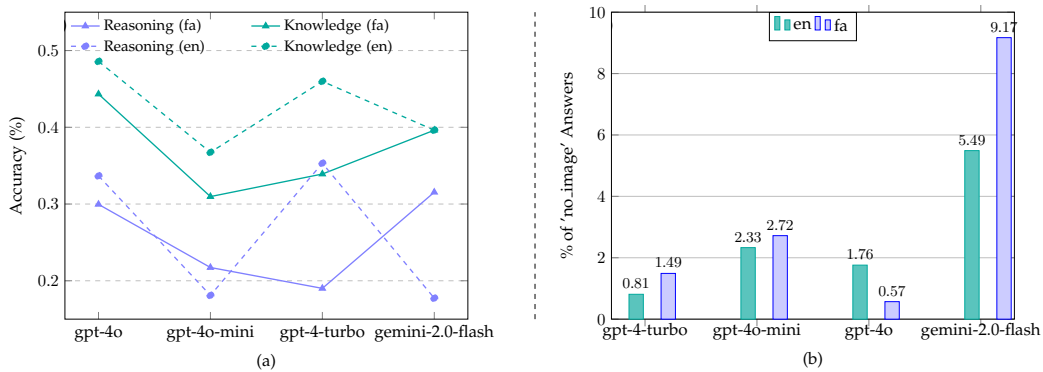

\begin{table}[!t]
\resizebox{\textwidth}{!}{
\begin{tabular}{|lccccc|}
\hline
\rule{0pt}{2ex}
            Methods\&Datasets     & \multicolumn{1}{c}{\textbf{Zero Shot}} & \multicolumn{1}{c}{\textbf{ICL}} & \multicolumn{1}{c}{\textbf{First Describe}} & \multicolumn{1}{c}{\textbf{Wrong Image}} & \multicolumn{1}{c|}{\textbf{Without Image}} \\
                 \hline
            
             \multicolumn{6}{l}{\rule{0pt}{2ex}\textbf{\textit{MEENA Persian Dataset}}}         \\
            \hline
            \rule{0pt}{2ex}
GPT-4o-mini      & 0.310 & 0.224 & 0.312 & 0.221 & 0.235 \\
GPT-4o           & 0.413 & \textbf{0.385} & 0.422 & \textbf{0.247} & \textbf{0.292} \\
GPT-4-Turbo      & 0.313 & 0.310 & 0.295 & - & 0.213 \\
Gemini-2.0-flash & \textbf{0.435} & 0.377 & \textbf{0.504} & 0.121 & 0.267 \\
\hline

        \multicolumn{6}{l}{\rule{0pt}{2ex}\textbf{\textit{MEENA English Dataset}}}  \\
                \hline
                 \rule{0pt}{2ex}
                 
GPT-4o-mini      & 0.368 & 0.312 & 0.361 & 0.275 & 0.279 \\
GPT-4o           & 0.474 & 0.397 & \textbf{0.464} & 0.269 & \textbf{0.401} \\
GPT-4-Turbo      & 0.440 & 0.384 & 0.381 & \textbf{0.306} & 0.304 \\
Gemini-2.0-flash & \textbf{0.494} & \textbf{0.464} & 0.459 & 0.178 & 0.311 \\
instructblip-t5  & 0.226 & * & 0.193 & 0.197 & *\\
\hline
                 \multicolumn{6}{l}{\rule{0pt}{2ex}\textbf{\textit{Art Persian Dataset}}}                                                                   \\
                 \hline
                 \rule{0pt}{2ex}
GPT-4o-mini      & 0.323 & 0.250 & 0.248 & \textbf{0.193} & 0.206 \\
GPT-4o           & \textbf{0.354} & 0.239 & 0.374 & 0.171 & 0.182 \\
GPT-4-Turbo      & 0.305 & 0.305 & 0.265 & - & 0.186 \\
Gemini-2.0-flash & 0.297 & \textbf{0.318} & \textbf{0.387} & 0.122 & \textbf{0.244} \\
\hline
\multicolumn{6}{l}{\rule{0pt}{2ex}\textbf{\textit{Art English Dataset}}} \\
\hline
\rule{0pt}{2ex}
GPT-4o-mini      & 0.343 & 0.276 & 0.301 & 0.241 & 0.217 \\
GPT-4o           & 0.372 & 0.311 & 0.406 & 0.230 & 0.232 \\
GPT-4-Turbo      & 0.336 & 0.374 & \textbf{0.334} & 0.197 & \textbf{0.294} \\
Gemini-2.0-flash & \textbf{0.376} & \textbf{0.372} & 0.329 & 0.151 & 0.159 \\
instructblip-t5  & 0.274 & * & 0.266 & \textbf{0.274} & * \\
\hline           
\end{tabular}
}
\caption{Accuracy comparison of different models across various tasks (Zero Shot, In-Context Learning, First Describe, Wrong Image, and Without Image) on the MEENA and Art datasets in both Persian and English. An asterisk (*) in the table indicates that the model does not support the corresponding setting.}
\end{table}
\noindent \textbf{Knowledge-Based Tasks Consistently Outperform Reasoning Ones}: The evaluation presented in Figure \ref{fig:grouped_plots}(a) highlights a performance gap between knowledge-based and reasoning tasks across various models. Knowledge-based tasks consistently outperform reasoning tasks by a significant margin of +10–19\% in absolute accuracy. This trend is observed for both English and Persian tasks, though Persian tasks generally exhibit lower accuracy, likely due to differences in training data distribution. These results suggest that while current vision-language models excel at factual recall, they face greater challenges with complex reasoning tasks. Moreover, the performance gap is more pronounced in Persian, indicating that reasoning tasks in this language are more difficult than in English.

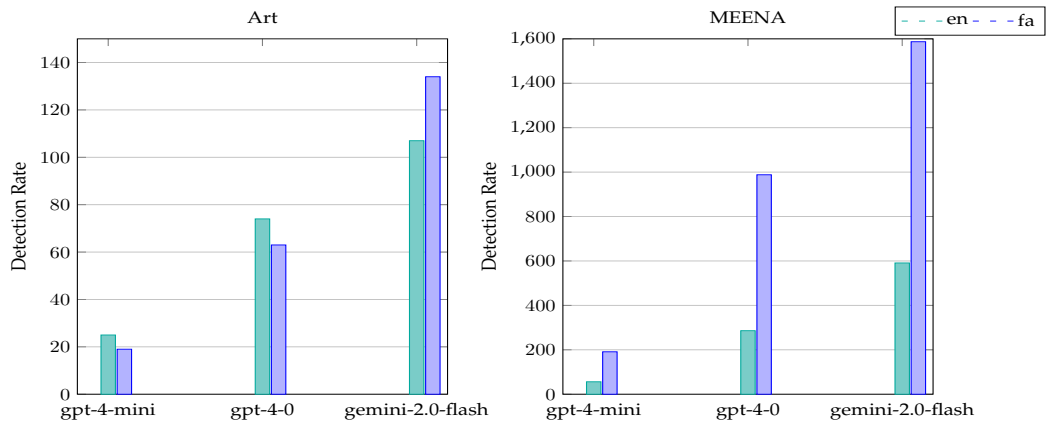
\begin{figure*}[h]
\resizebox{\textwidth}{0.4\textwidth}{
    \begin{tikzpicture}
	\begin{groupplot}[
		group style={
			group size=2 by 1, 
			horizontal sep=2cm,
			vertical sep=0.5cm,
		},
		width=8cm,
		height=9cm,
		ylabel={Detection Rate},
		symbolic x coords={gpt-4-mini, gpt-4-0, gemini-2.0-flash},
		xtick=data,
		ymajorgrids=true,
		x tick label style={rotate=0, anchor=center, yshift=-10pt},
		legend style={at={(1.1,1.05)},anchor=center,legend columns=-1},
		]
		
		\nextgroupplot[title={Art}, ymin=0, ymax=150]
		\addplot[ybar, draw=Emerald, fill=Emerald!50, bar width=0.25 cm] coordinates { (gpt-4-mini, 25) (gpt-4-0, 74) (gemini-2.0-flash, 107)};
		\addplot[ybar, draw=blue, fill=blue!30, bar width=0.25 cm,  xshift=0.28cm] coordinates {(gpt-4-mini, 19) (gpt-4-0, 63) (gemini-2.0-flash, 134)};
		
		\nextgroupplot[title={MEENA}, ymin=0, ymax=1600]
		\addplot[ybar, draw=Emerald, fill=Emerald!50, bar width=0.25 cm] coordinates { (gpt-4-mini, 56) (gpt-4-0, 286) (gemini-2.0-flash, 591)};
		\addplot[ybar, draw=blue, fill=blue!30, bar width=0.25 cm, xshift=0.28cm] coordinates { (gpt-4-mini, 191) (gpt-4-0, 988) (gemini-2.0-flash, 1587)};
		
		\legend{en, fa}
		
	\end{groupplot}
\end{tikzpicture}
}
\caption{Hallucination detection rates across three vision-language models (GPT-4 Mini, GPT-4, and Gemini 2.0 Flash) on the Art and MEENA datasets, for both English and Persian. The bars represent detection rates for each model in both languages, with a clear performance gap observed in Persian, particularly for Gemini 2.0 Flash.}
\label{fig:hallucination}
\end{figure*}

\noindent \textbf{Hallucination Detection Performance with Incorrect Images}: Figure \ref{fig:hallucination} compares hallucination detection rates across three vision-language models—Gemini 2.0 Flash, GPT-4, and GPT-4 Mini—on the Art and MEENA datasets in both English and Persian. To evaluate hallucination detection, we replace each query's image with an incorrect one (Section \ref{ss:wrong_image}) and consider a detection successful only if the model identifies the mismatch.
Gemini 2.0 Flash consistently achieves higher detection rates than both GPT-4 and GPT-4 Mini across datasets, with a particularly significant performance gap in Persian. The detection rate difference between Gemini 2.0 Flash and GPT-4 Mini on the MEENA dataset is over 400 detections, suggesting that Gemini 2.0 Flash is more robust at recognizing inconsistencies, especially in Persian contexts.

\noindent \textbf{\textit{No Image} Errors in Image Detection Across Models}: Figure \ref{fig:grouped_plots}(b) illustrates the frequency at which different models mistakenly report the absence of an image, despite one being provided. The chart displays the percentage of 'no image' responses for four models, evaluated on both English and Persian inputs. GPT-4-Turbo and GPT-4o demonstrate relatively low 'no image' error rates across both languages, with English inputs yielding slightly fewer errors than Persian. In contrast, Gemini 2.0 Flash exhibits a markedly higher incidence of 'no image' responses, particularly for Persian inputs, where the error rate reaches 9.17\%.

\noindent \textbf{Models Struggle with Higher-Level Questions}: Figures \ref{fig:Heatmap-chemistry-zero-shot-MEENA-en} and \ref{fig:Heatmap-math-zero-shot-MEENA-en} show that as question difficulty increases in the Chemistry and Mathematics tasks of the zero-shot experiment in English, model performance generally declines. While models like GPT-4o-mini and GPT-4-Turbo experience significant drops in accuracy at higher levels, Gemini-2.0-flash maintains relatively consistent performance, particularly in the Mathematics task. In contrast, instructblip-t5 struggles across all levels, especially in the Chemistry task. Further results are provided in the appendix \ref{app:results}.

\begin{figure*}[h]
    \hspace*{-2em}
    \pgfplotstabletypeset[
	color cells={min=0,max=1},
	col sep=comma,
	/pgfplots/colormap={whitegreen}{rgb255(0cm)=(255,255,255); rgb255(1cm)=(80,200,204)},
	columns/Model/.style={reset styles,string type}, font=\small
	]{
		Model,lvl 12,lvl 11,lvl 10,lvl 9,lvl 8,lvl 7,lvl 6,lvl 5,lvl 4,lvl 3,lvl 2
		GPT-4o-mini,0.29,	0.27,	0.32,	0.5,	0.49,	0.57,	0.33,	0.65,	0.57,	0.57,	0.71

		GPT-4o, 0.38,	0.37,	0.48,	0.64,	0.61,	0.68,	0.63,	0.75,	0.77,	0.72,	0.87
        
		GPT-4-Turbo,0.23,	0.32,	0.48,	0.57,	0.56,	0.68,	0.52,	0.68,	0.52,	0.6,	0.81

		Gemini-2.0-flash,0.47,	0.40,	0.72,	0.59,	0.72,	0.75,	0.59,	0.68,	0.66,	0.62,	0.76

		instructblip-t5,0.42,	0.16,	0.2,	0.36,	0.28,	0.27,	0.14,	0.42,	0.34,	0.37,	0.42
	}
    \caption{Heatmap of model accuracy across different levels of the \textbf{MEENA English} dataset for the \textbf{Chemistry Course/Experimental Science} in the \textbf{Zero-shot} experiment.}
    \label{fig:Heatmap-chemistry-zero-shot-MEENA-en}
\end{figure*}

\begin{figure*}[h]
    \hspace*{-2em}
    \pgfplotstabletypeset[
	color cells={min=0,max=1},
	col sep=comma,
	/pgfplots/colormap={whitegreen}{rgb255(0cm)=(255,255,255); rgb255(1cm)=(80,200,204)},
	columns/Model/.style={reset styles,string type}, 
        font=\small
	]{
		Model,lvl 12,lvl 11,lvl 10,lvl 9,lvl 8,lvl 7,lvl 6,lvl 5,lvl 4,lvl 3,lvl 2
		GPT-4o-mini,0.45,	0.23,	0.41,	0.44,	0.29,	0.35,	0.46,	0.33,	0.31,	0.42,	0.36

		GPT-4o, 0.35,	0.41,	0.5,	0.48,	0.37,	0.64,	0.37,	0.55,	0.65,	0.5,	0.60
        
		GPT-4-Turbo,0.25,	0.32,	0.38,	0.44,	0.34,	0.47,	0.48,	0.5,	0.42,	0.5,	0.49

		Gemini-2.0-flash,0.48,	0.57,	0.53,	0.63,	0.53,	0.64,	0.43,	0.5,	0.47,	0.58,	0.57

		instructblip-t5,0.16,	0.27,	0.28,	0.14,	0.16,	0.20,	0.2,	0.22,	0.19,	0.14,	0.21
	}
    \caption{Heatmap of model accuracy across different levels of the \textbf{MEENA English} dataset for the \textbf{Mathematics} in the \textbf{Zero-Shot} experiment.}
    \label{fig:Heatmap-math-zero-shot-MEENA-en}
\end{figure*}

\section{Conclusions}
\label{sec:conclusions}
In this study, we present MEENA, the first benchmark designed to assess scientific reasoning, problem-solving, and human-level Persian language understanding in VLMs. MEENA comprises multiple-choice questions available in both Persian and English, enriched with extensive metadata, including difficulty levels and descriptive answers. Furthermore, we conducted a series of experiments to analyze different model capabilities, including Zero-Shot, In-Context Learning, First Describe, Wrong Image, and Without Image scenarios. Our evaluation highlights key performance trends across vision-language models. (1) Knowledge-based tasks consistently outperform reasoning-based ones, with a more pronounced gap in Persian. (2) Gemini 2.0-flash surpasses GPT-4 and GPT-4o-Mini in detecting image mismatches, demonstrating greater reliability in mitigating hallucinations, particularly in Persian. (3) GPT-4-Turbo and GPT-4o excel in image presence detection, while Gemini 2.0-flash shows higher 'no image' error rates. (4) Models struggle with higher-level Chemistry and Mathematics questions, with performance declining as complexity increases. These findings emphasize the challenges of complex reasoning and domain-specific knowledge retrieval in both Persian and English for vision-language models.




\clearpage
\bibliography{colm2025_conference}
\bibliographystyle{colm2025_conference}

\clearpage
\appendix

\section{Dataset Additional Information}
\label{app:dataset}

We provide several examples from our dataset, MEENA, showcasing the diversity and structure of the questions included. Each example contains a visual component along with corresponding questions in both Persian and English. These samples illustrate different question formats, such as multiple-choice questions, mathematical problem-solving, and pattern recognition, all integrated with images. 

\begin{figure}[H]
	\centering
	\begin{tabular}{cc}
		\resizebox{5cm}{!}{
			\begin{tcolorbox}[colframe=Emerald, colback=blue!5, width=0.6\textwidth]
				\includegraphics[width=\textwidth]{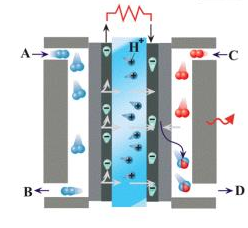}\\
				\textbf{Persian question:} \\
				
				\begin{otherlanguage}{farsi}\small
					شکل مقابل نشان‌دهنده رایج‌ترین سلول سوختی است. چند مورد از مطالب زیر نادرست هستند؟ 
					\vskip 0.2cm
					آ) در این سلول دو گاز به‌طور کنترل‌شده با یکدیگر وارد واکنش می‌شوند و در حدود ۶۰ درصد از انرژی شیمیایی تولیدی به انرژی الکتریکی تبدیل می‌شود. 
					
					\vskip 0.2cm
					ب) واکنش کلی انجام‌شده در این سلول به‌صورت $2{{H}_{2}}(g)+{{O}_{2}}(g)\to 2{{H}_{2}}O\,(l)$است. 
					
					\vskip 0.2cm
					پ) در این سلول جریان الکترون‌ها در مدار بیرونی برخلاف جریان پروتون‌ها در غشای مبادله‌کننده پروتون، از آند به کاتد است. 
					
					\vskip 0.2cm
					ت) گاز $B$ همان گاز $A$ است که می‌تواند به عنوان سوخت این سلول به‌طور پیوسته وارد سلول شده و اکسایش یابد.
				\end{otherlanguage}		
				
				$(C=12\,,\,O=16\,\,,\,\,H=1:g.mo{{l}^{-1}})$
				\begin{otherlanguage}{farsi}\small
					
					1-
					۱
					
					2-
					۲
					
					3-
					۳
					
					4-
					۴
					
				\end{otherlanguage}
				
				\vspace{0.5cm}
				\hrule
				\vspace{0.5cm}
				
				\textbf{English question:} \\
				\small
				The figure opposite represents the most common fuel cell. How many of the following statements are incorrect?
				
				A) In this cell, two gases react with each other in a controlled manner, and about 60\% of the generated chemical energy is converted into electrical energy.
				
				B) The overall reaction occurring in this cell is represented as $2{{H}_{2}}(g)+{{O}_{2}}(g)\to 2{{H}_{2}}O\,(l)$.
				
				C) In this cell, the flow of electrons in the external circuit is from anode to cathode, opposite to the flow of protons in the proton exchange membrane.
				
				D) Gas $B$ is the same as gas $A$, which can continuously enter the cell as fuel and be oxidized.
				
				1) 1
				2) 2
				3) 3
				4) 4
			\end{tcolorbox}
		}
		
		&
		
		\resizebox{5cm}{!}{
			\begin{tcolorbox}[colframe=Emerald, colback=blue!5, width=0.6\textwidth]
				\includegraphics[width=\textwidth]{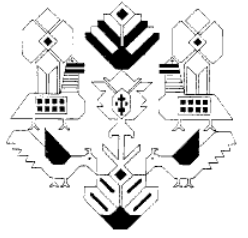}\\
                \vskip 1cm
				\textbf{Persian question:} \\
				
				\begin{otherlanguage}{farsi}\small
					طرح رو به رو با موضوع مرغ و درخت، از دست‌بافت‌های سنتی کدام اقوام است؟
					\vskip 0.2cm
	
				\end{otherlanguage}		
                
				\begin{otherlanguage}{farsi}\small
					\vskip 0.5cm
									1- ایل عرب - خوزستان
					
					\vskip 0.75cm
					2- ایل بهارلو - فارس
					
					\vskip 0.75cm
					3- لک - کرمانشاه 
					
					\vskip 0.75cm
					4- ترک - همدان
					
				\end{otherlanguage}
				
				\vspace{0.5cm}
				\hrule
				\vspace{0.5cm}

                \vskip 0.5cm
				\textbf{English question:} \\
				\small
				The design featuring a chicken and a tree, shown in front, is a traditional handmade craft of which ethnic group?
                
				\vskip 0.5cm
				1) Arab Tribe - Khuzestan

                \vskip 0.5cm
                2) Baharlu Tribe - Fars

                \vskip 0.5cm
                3) Lak - Kermanshah
                
                \vskip 0.5cm
                4) Turk - Hamedan
                
			\end{tcolorbox}
		}
	\end{tabular}
	\caption{Sample of MEENA questions}
	\label{fig:question_layout}
\end{figure}

\begin{figure}[H]
	\centering
	\begin{tabular}{cc}
		\resizebox{5cm}{!}{
			\begin{tcolorbox}[colframe=Emerald, colback=blue!5, width=0.6\textwidth]
				
				\textbf{Question} \\
				
				\begin{otherlanguage}{farsi}\small
                مجموع انگشتان باز دست راست و انگشتان باز دست چپ شکل زیر، در کدام گزینه به صورت صحیح آمده است؟
				\end{otherlanguage}		
                
                \textit{Which option correctly states the sum of the open fingers of the right hand and the open fingers of the left hand in the figure below?}

                \vskip 0.1cm
				\includegraphics[width=0.6\textwidth]{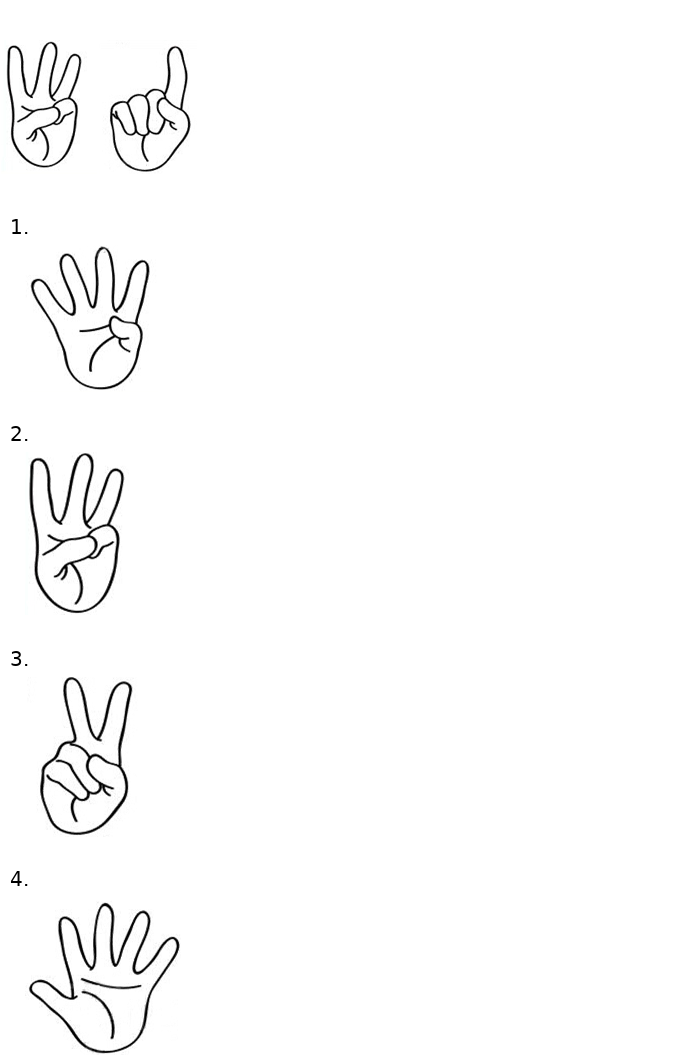}
				
			\end{tcolorbox}
		}
		
		&
		
		\resizebox{5cm}{!}{
			\begin{tcolorbox}[colframe=Emerald, colback=blue!5, width=0.6\textwidth]
				\textbf{Question:} \\
				
				\begin{otherlanguage}{farsi}\small
ا ز کدام شکل گسترده مکعبی با نمای روبه‌رو حاصل می‌شود؟ پشت برگه‌ها کاملاً سفید است.				
\vskip 0.2cm
	
				\end{otherlanguage}		

                \vskip 0.3cm
                \textit{From which unfolded shape is the cube with the front view obtained? The back of the sheets is completely white.}
            \\
				
			\includegraphics[width=0.6\textwidth]{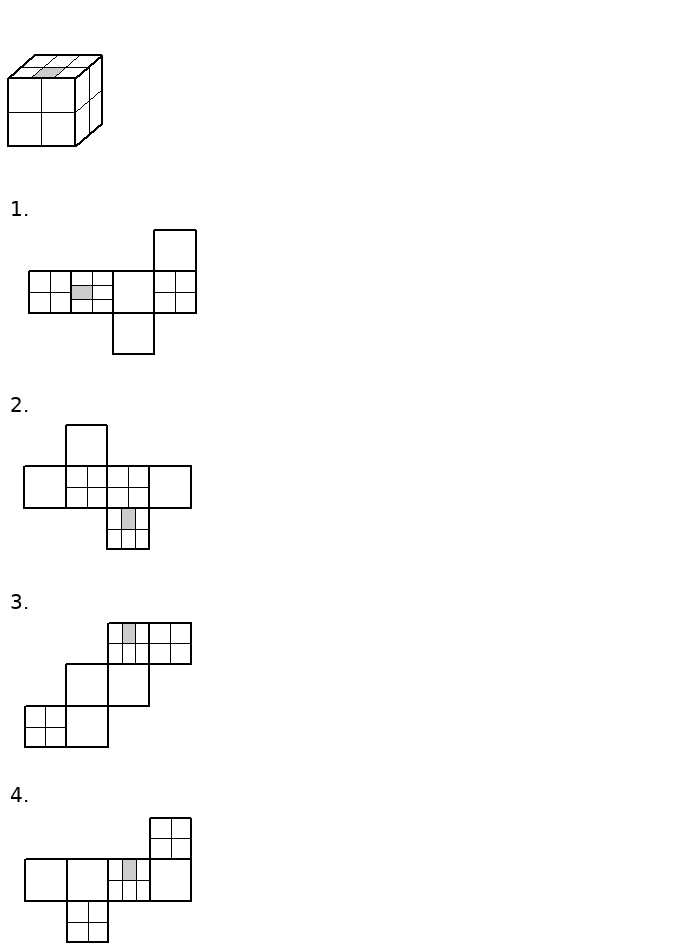}
			\end{tcolorbox}
		}

        \\

        \resizebox{5cm}{!}{
			\begin{tcolorbox}[colframe=Emerald, colback=blue!5, width=0.6\textwidth]
				
				\textbf{Question} \\
				
				\begin{otherlanguage}{farsi}\small
                عدد احاطه‌گری کدام گزینه متفاوت از گزینه‌های دیگر است؟				          
                    \end{otherlanguage}		

                \vskip 0.1cm
				\includegraphics[width=\textwidth]{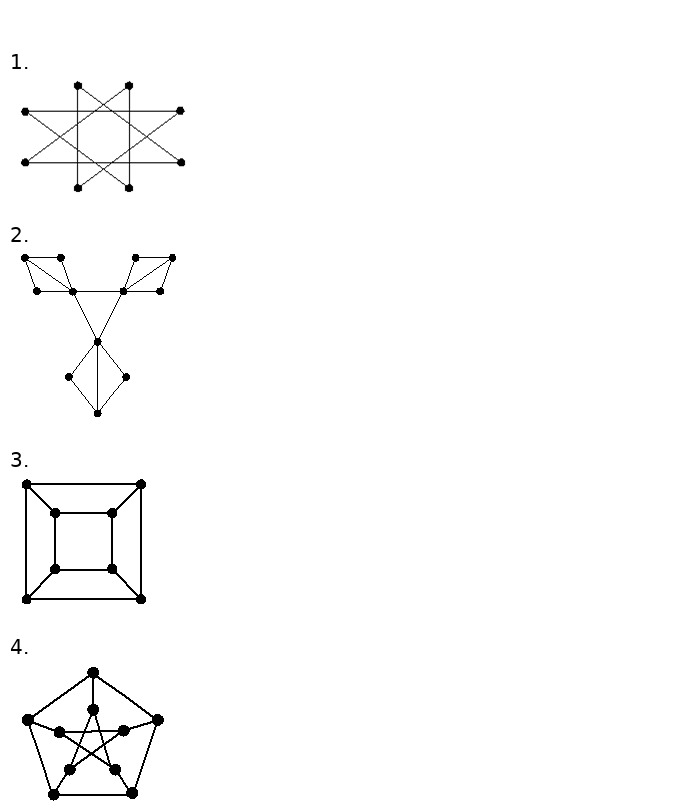}
				
			\end{tcolorbox}
		}
		
		&
		
		\resizebox{5cm}{!}{
			\begin{tcolorbox}[colframe=Emerald, colback=blue!5, width=0.6\textwidth]
				\textbf{Question:} \\
				
				\begin{otherlanguage}{farsi}\small
غذای کدام جانور، میوه یا دانه نیست ؟					\vskip 0.2cm
	
				\end{otherlanguage}		

                \vskip 0.3cm
                \textit{Which animal's food is not fruit or seeds?}
            \\
				
			\includegraphics[width=\textwidth]{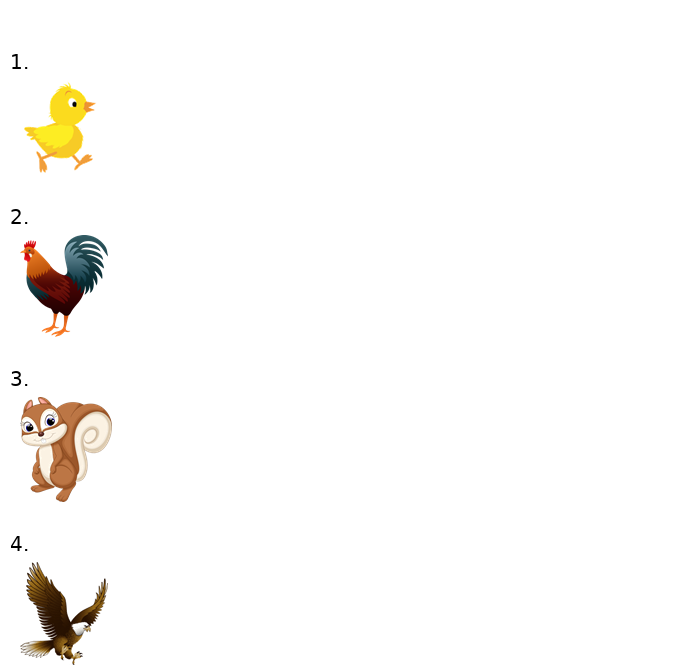}
			\end{tcolorbox}
		}
	\end{tabular}
	\caption{Sample of MEENA questions with picture in choices}
	\label{fig:question_layout2}
\end{figure}

\begin{figure}[H]
	\centering
	\begin{tabular}{cc}
		\resizebox{5cm}{!}{
			\begin{tcolorbox}[colframe=Emerald, colback=blue!5, width=0.6\textwidth]
				\includegraphics[width=\textwidth]{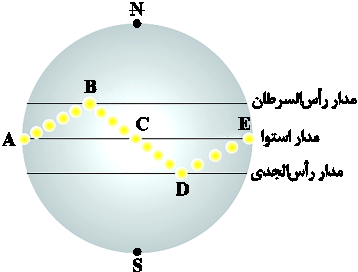}\\
				\textbf{Persian question:} \\
				
				\begin{otherlanguage}{farsi}\small
                اگر خط $g$ مطابق شکل زیر در نقطه $A(2,3)$ بر نمودار$f(x)$ مماس و ${f}`(2)=2$ باشد، آن‌گاه عرض از مبدأ خط $g$ کدام است؟
                
				\end{otherlanguage}		
                \vskip 1cm
				\begin{otherlanguage}{farsi}\small
					
					1-
					1-
					\vskip 0.5cm
                    
					2-
					2
					\vskip 0.5cm
                    
					3-
					1
                        \vskip 0.5cm
					
					4-
					3-
					
				\end{otherlanguage}
				
			\end{tcolorbox}
		}
		
		&
		
		\resizebox{5cm}{!}{
			\begin{tcolorbox}[colframe=Emerald, colback=blue!5, width=0.6\textwidth]
				\includegraphics[width=\textwidth]{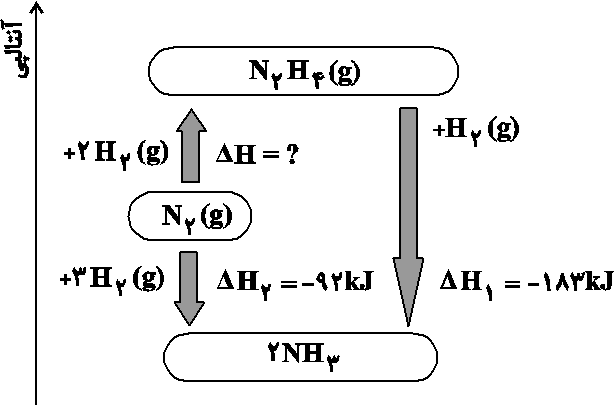}\\
                \vskip 0.1cm
				\textbf{Persian question:} \\
				
				\begin{otherlanguage}{farsi}\small
				با توجه به نمودار زیر برای تولید$160$ گرم هیدرازین از گازهای نیتروژن و هیدروژن، چند کیلوژول انرژی لازم است؟ 
 
					\vskip 0.2cm
	
				\end{otherlanguage}		

                $(H=1\,\,,\,\,\,\,N=14:g\,.\,mo{{l}^{-1}})$
				\begin{otherlanguage}{farsi}\small
					\vskip 0.5cm
									1- 5/225
					
					\vskip 0.75cm
2- 554					
					\vskip 0.75cm
					3- 019 
					
					\vskip 0.75cm
					4- 5731
					
				\end{otherlanguage}
				 
			\end{tcolorbox}
		}
	\end{tabular}
	\caption{Sample of MEENA questions including Persian texts in picture}
	\label{fig:question_layout3}
\end{figure}

\section{Experiments Additional Information}
\label{app:experiment}

\subsection{Models Used}

We evaluate the following models in our experiments:

\begin{itemize}
    \item \textbf{GPT-4o} and \textbf{GPT-4o-mini}: Larger and smaller variants of OpenAI's GPT-4-based architecture capable of processing text, images, and audio, designed for real-time multimodal interaction \citep{openai2024gpt4o}.
    
    \item \textbf{GPT-4-Turbo}: An optimized variant of GPT-4, developed by OpenAI, suited for interactive dialogue with improved cost and performance characteristics \citep{openai2023gpt4turbo}.
    
    \item \textbf{Gemini-2.0-flash}: A multimodal vision-language model developed by Google DeepMind, trained to process and integrate text, image, and video inputs efficiently \citep{google2023gemini}.
    
    \item \textbf{InstructBLIP-T5}: A T5-based vision-language model that incorporates instruction tuning and visual grounding to handle complex multimodal tasks \citep{dai2023instructblip}.
\end{itemize}

By evaluating all models on the same tasks and under each of the five experimental settings, we can measure their relative strengths and weaknesses in multimodal reasoning.

\subsection{Experimental Cases and Motivations}
We apply each of the five experiment types (ZS, ICL, FD, WI, WO) to the three image-based question categories introduced earlier: (1) questions with images, (2) choices with images, and (3) both questions and choices containing images. The rationale for each experiment type is as follows:
\begin{itemize}
    \item \textbf{Zero-Shot (ZS)}: Establishes a baseline for model performance without contextual examples.
    \item \textbf{In-Context Learning (ICL)}: Investigates whether a few-shot prompt improves multimodal understanding.
    \item \textbf{First Describe (FD)}: Tests whether forcing a detailed image description yields more accurate reasoning.
    \item \textbf{Wrong Image (WI)}: Assesses how reliant the model is on correct image cues (detecting mismatches, etc.).
    \item \textbf{Without Image (WO)}: Shows performance under pure text-only conditions, contrasting it with results that use images.
\end{itemize}

\section{Results}
\label{app:results}
\begin{figure}[H]
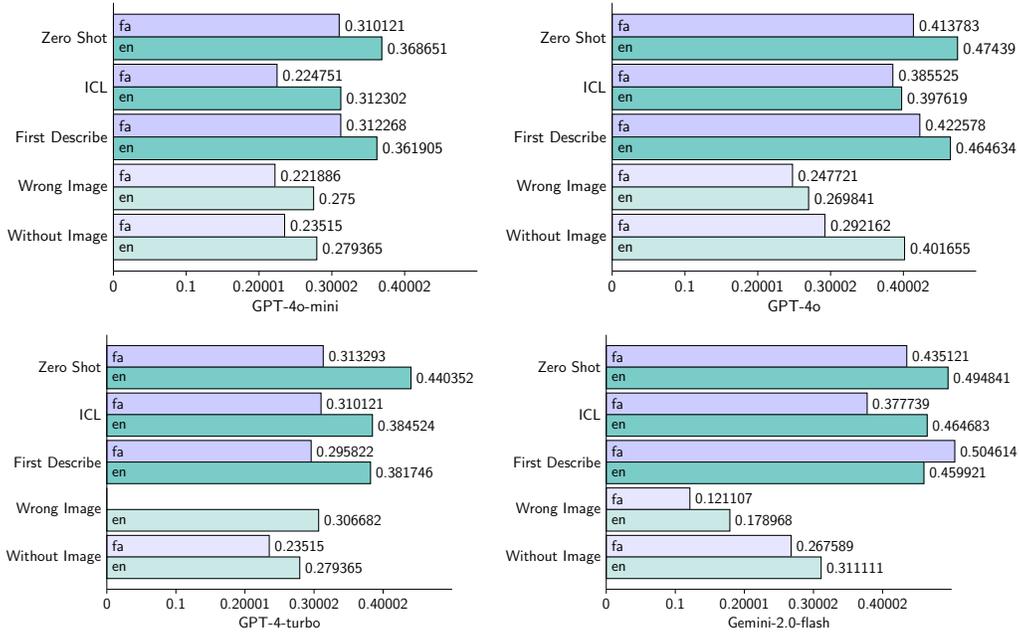

    \begin{center}
        \resizebox{\textwidth}{!}{
    \begin{tabular}{cc}
        \begin{bchart}[step=0.1,max=0.5]
            \bcbar[text=fa]{0.310121}
            \bclabel{Zero Shot}
            \bcbar[text=en, color=Emerald!50]{0.368651}
            \bcskip{0.1cm}
            \bcbar[text=fa]{0.224751}
            \bclabel{ICL}
            \bcbar[text=en, color=Emerald!50]{0.312302}
            \bcskip{0.1cm}
            \bcbar[text=fa]{0.312268}
            \bclabel{First Describe}
            \bcbar[text=en, color=Emerald!50]{0.361905}
            \bcskip{0.1cm}
            \bcbar[text=fa, color=blue!10]{0.221886}
            \bclabel{Wrong Image}
            \bcbar[text=en, color=Emerald!20]{0.275}
            \bcskip{0.1cm}
            \bcbar[text=fa, color=blue!10]{0.23515}
            \bclabel{Without Image}
            \bcbar[text=en,color=Emerald!20]{0.279365}
            \bcxlabel{GPT-4o-mini}
        \end{bchart} &  
        \begin{bchart}[step=0.1,max=0.5]
            \bcbar[text=fa]{0.413783}
            \bclabel{Zero Shot}
            \bcbar[text=en, color=Emerald!50]{0.47439}
            \bcskip{0.1cm}
            
            \bcbar[text=fa]{0.385525}
            \bclabel{ICL}
            \bcbar[text=en, color=Emerald!50]{0.397619}
            \bcskip{0.1cm}
            
            \bcbar[text=fa]{0.422578}
            \bclabel{First Describe}
            \bcbar[text=en, color=Emerald!50]{0.464634}
            \bcskip{0.1cm}
            
            \bcbar[text=fa, color=blue!10]{0.247721}
            \bclabel{Wrong Image}
            \bcbar[text=en, color=Emerald!20]{0.269841}
            \bcskip{0.1cm}
            
            \bcbar[text=fa, color=blue!10]{0.292162}
            \bclabel{Without Image}
            \bcbar[text=en, color=Emerald!20]{0.401655}
            \bcxlabel{GPT-4o}
        \end{bchart}  
    \end{tabular}
    }
    \end{center}
    
    \begin{center}
    \resizebox{\textwidth}{!}{
    \begin{tabular}{cc} 
        \begin{bchart}[step=0.1,max=0.5]
            \bcbar[text=fa]{0.313293}
            \bclabel{Zero Shot}
            \bcbar[text=en, color=Emerald!50]{0.440352}
            \bcskip{0.1cm}
            
            \bcbar[text=fa]{0.310121}
            \bclabel{ICL}
            \bcbar[text=en, color=Emerald!50]{0.384524}
            \bcskip{0.1cm}
            
            \bcbar[text=fa]{0.295822}
            \bclabel{First Describe}
            \bcbar[text=en, color=Emerald!50]{0.381746}
            \bcskip{0.1cm}
            
            \bcbar[text=]{}
            \bclabel{Wrong Image}
            \bcbar[text=en, color=Emerald!20]{0.306682}
            \bcskip{0.1cm}
            
            \bcbar[text=fa, color=blue!10]{0.23515}
            \bclabel{Without Image}
            \bcbar[text=en, color=Emerald!20]{0.279365}
            \bcxlabel{GPT-4-turbo}
        \end{bchart}  &
        \begin{bchart}[step=0.1,max=0.5]
            \bcbar[text=fa]{0.435121}
            \bclabel{Zero Shot}
            \bcbar[text=en, color=Emerald!50]{0.494841}
            \bcskip{0.1cm}
            
            \bcbar[text=fa]{0.377739}
            \bclabel{ICL}
            \bcbar[text=en, color=Emerald!50]{0.464683}
            \bcskip{0.1cm}
            
            \bcbar[text=fa]{0.504614}
            \bclabel{First Describe}
            \bcbar[text=en, color=Emerald!50]{0.459921}
            \bcskip{0.1cm}
            
            \bcbar[text=fa, color=blue!10]{0.121107}
            \bclabel{Wrong Image}
            \bcbar[text=en, color=Emerald!20]{0.178968}
            \bcskip{0.1cm}
            
            \bcbar[text=fa, color=blue!10]{0.267589}
            \bclabel{Without Image}
            \bcbar[text=en, color=Emerald!20]{0.311111}
            \bcxlabel{Gemini-2.0-flash}
        \end{bchart}
         \\
    \end{tabular}
    }    
    \end{center}
    
    \caption{Comparison of Farsi and English performance across different experiments and models on the \textbf{MEENA} dataset.}
    \label{fig:Fa_En_Comparison_MEENA}
\end{figure}

\begin{figure}[H]
    \begin{center}
        \resizebox{\textwidth}{!}{
    \begin{tabular}{cc}
        \begin{bchart}[step=0.1,max=0.5]
            \bcbar[text=fa]{0.323583}
            \bclabel{Zero Shot}
            \bcbar[text=en, color=Emerald!50]{0.343693}
            \bcskip{0.1cm}
            \bcbar[text=fa]{0.250457}
            \bclabel{ICL}
            \bcbar[text=en, color=Emerald!50]{0.276051}
            \bcskip{0.1cm}
            \bcbar[text=fa]{0.248629}
            \bclabel{First Describe}
            \bcbar[text=en, color=Emerald!50]{0.301645}
            \bcskip{0.1cm}
            \bcbar[text=fa, color=blue!10]{0.193784}
            \bclabel{Wrong Image}
            \bcbar[text=en, color=Emerald!20]{0.241316}
            \bcskip{0.1cm}
            \bcbar[text=fa, color=blue!10]{0.206581}
            \bclabel{Without Image}
            \bcbar[text=en,color=Emerald!20]{0.21755}
            \bcxlabel{GPT-4o-mini}
        \end{bchart} &  
        \begin{bchart}[step=0.1,max=0.5]
            \bcbar[text=fa]{0.354662}
            \bclabel{Zero Shot}
            \bcbar[text=en, color=Emerald!50]{0.372943}
            \bcskip{0.1cm}
            
            \bcbar[text=fa]{0.239488}
            \bclabel{ICL}
            \bcbar[text=en, color=Emerald!50]{0.310786}
            \bcskip{0.1cm}
            
            \bcbar[text=fa]{0.374771}
            \bclabel{First Describe}
            \bcbar[text=en, color=Emerald!50]{0.40585}
            \bcskip{0.1cm}
            
            \bcbar[text=fa, color=blue!10]{0.171846}
            \bclabel{Wrong Image}
            \bcbar[text=en, color=Emerald!20]{0.230347}
            \bcskip{0.1cm}
            
            \bcbar[text=fa, color=blue!10]{0.182815}
            \bclabel{Without Image}
            \bcbar[text=en, color=Emerald!20]{0.232176}
            \bcxlabel{GPT-4o}
        \end{bchart} 
    \end{tabular}
    }
    \end{center}
    
    \begin{center}
    \resizebox{\textwidth}{!}{
    \begin{tabular}{cc}
        \begin{bchart}[step=0.1,max=0.5]
            \bcbar[text=fa]{0.305302}
            \bclabel{Zero Shot}
            \bcbar[text=en, color=Emerald!50]{0.33638}
            \bcskip{0.1cm}
            
            \bcbar[text=fa]{0.305302}
            \bclabel{ICL}
            \bcbar[text=en, color=Emerald!50]{0.374771}
            \bcskip{0.1cm}
            
            \bcbar[text=fa]{0.265082}
            \bclabel{First Describe}
            \bcbar[text=en, color=Emerald!50]{0.334552}
            \bcskip{0.1cm}
            
            \bcbar[text=fa, color=blue!10]{0.184644}
            \bclabel{Wrong Image}
            \bcbar[text=en, color=Emerald!20]{0.197441}
            \bcskip{0.1cm}
            
            \bcbar[text=fa, color=blue!10]{0.186472}
            \bclabel{Without Image}
            \bcbar[text=en, color=Emerald!20]{0.294333}
            \bcxlabel{GPT-4-turbo}
        \end{bchart}  
        &
         \begin{bchart}[step=0.1,max=0.5]
            \bcbar[text=fa]{0.297989}
            \bclabel{Zero Shot}
            \bcbar[text=en, color=Emerald!50]{0.3766}
            \bcskip{0.1cm}
            
            \bcbar[text=fa]{0.318099}
            \bclabel{ICL}
            \bcbar[text=en, color=Emerald!50]{0.372943}
            \bcskip{0.1cm}
            
            \bcbar[text=fa]{0.387569}
            \bclabel{First Describe}
            \bcbar[text=en, color=Emerald!50]{0.329068}
            \bcskip{0.1cm}
            
            \bcbar[text=fa, color=blue!10]{0.122486}
            \bclabel{Wrong Image}
            \bcbar[text=en, color=Emerald!20]{0.151737}
            \bcskip{0.1cm}
            
            \bcbar[text=fa, color=blue!10]{0.244973}
            \bclabel{Without Image}
            \bcbar[text=en, color=Emerald!20]{0.159049}
            \bcxlabel{Gemini-2.0-flash}
        \end{bchart} 
    \end{tabular}
    }    
    \end{center}
    
    \caption{Comparison of Farsi and English performance across different experiments and models on the \textbf{Art} dataset.}
    \label{fig:Fa_En_Comparison_MEENA}
\end{figure}

\begin{figure}[H]
\resizebox{\textwidth}{!}{
    \begin{tabular}{ccc}
    \begin{bchart}[step=0.1,max=0.5]
        \bcbar[text=GPT-4o-mini, color=BlueViolet!20]{0.368151}
        \bcskip{0.1cm}
        \bcbar[text=GPT-4o, color=CornflowerBlue!20]{0.4617}
        \bcskip{0.1cm}
        \bcbar[text=GPT-4-turbo, color=Fuchsia!20]{0.47439}
        \bcskip{0.1cm}
        \bcbar[text=Gemini 2.0 flash, color=SpringGreen!20]{0.494841}
        \bcskip{0.1cm}
        \bcbar[text=instructblip-t5, color=RawSienna!20]{0.226984}
        \bcskip{0.1cm}
        \bcxlabel{Zero Shot (en)}
    \end{bchart}&  
        \begin{bchart}[step=0.1,max=0.5]
        \bcbar[text=GPT-4o-mini, color=BlueViolet!20]{0.312302}
        \bcskip{0.1cm}
        \bcbar[text=GPT-4o, color=CornflowerBlue!20]{0.397619}
        \bcskip{0.1cm}
        \bcbar[text=GPT-4-turbo, color=Fuchsia!20]{0.384524}
        \bcskip{0.1cm}
        \bcbar[text=Gemini 2.0 flash, color=SpringGreen!20]{0.464683}
        \bcskip{0.1cm}
        \bcbar[text=instructblip-t5, color=RawSienna!20]{}
        \bcskip{0.1cm}
            \bcxlabel{ICL (en)}
        \end{bchart} &
        \begin{bchart}[step=0.1,max=0.5]
        \bcbar[text=GPT-4o-mini, color=BlueViolet!20]{0.361905}
        \bcskip{0.1cm}
        \bcbar[text=GPT-4o, color=CornflowerBlue!20]{0.464634}
        \bcskip{0.1cm}
        \bcbar[text=GPT-4-turbo, color=Fuchsia!20]{0.381746}
        \bcskip{0.1cm}
        \bcbar[text=Gemini 2.0 flash, color=SpringGreen!20]{0.459921}
        \bcskip{0.1cm}
        \bcbar[text=instructblip-t5, color=RawSienna!20]{0.193651}
        \bcskip{0.1cm}
            \bcxlabel{First Describe (en)}
        \end{bchart}  \\
    \end{tabular}
}

\begin{center}
    \resizebox{10cm}{!}{
        \begin{tabular}{cc}
            \begin{bchart}[step=0.1,max=0.5]
            \bcbar[text=GPT-4o-mini, color=BlueViolet!20]{0.275}
            \bcskip{0.1cm}
            \bcbar[text=GPT-4o, color=CornflowerBlue!20]{0.269841}
            \bcskip{0.1cm}
            \bcbar[text=GPT-4-turbo, color=Fuchsia!20]{0.381746}
            \bcskip{0.1cm}
            \bcbar[text=Gemini 2.0 flash, color=SpringGreen!20]{0.178968}
            \bcskip{0.1cm}
            \bcbar[text=instructblip-t5, color=RawSienna!20]{0.197222}
            \bcskip{0.1cm}
                \bcxlabel{Wrong Image (en)}
            \end{bchart} &  
            \begin{bchart}[step=0.1,max=0.5]
            \bcbar[text=GPT-4o-mini, color=BlueViolet!20]{0.279365}
            \bcskip{0.1cm}
            \bcbar[text=GPT-4o, color=CornflowerBlue!20]{0.401655}
            \bcskip{0.1cm}
            \bcbar[text=GPT-4-turbo, color=Fuchsia!20]{0.304762}
            \bcskip{0.1cm}
            \bcbar[text=Gemini 2.0 flash, color=SpringGreen!20]{0.311111}
            \bcskip{0.1cm}
            \bcbar[text=instructblip-t5, color=RawSienna!20]{}
            \bcskip{0.1cm}
                \bcxlabel{Without Image (en)}
            \end{bchart} 
        \end{tabular}
    }
\end{center}
    \caption{Performance comparison of each model across experiments on the \textbf{MEENA English} dataset}
    \label{fig:BarChart-MEENA-En-AllModelComparison}
\end{figure}
\begin{figure}[H]
\resizebox{\textwidth}{!}{
    \begin{tabular}{ccc}
    \begin{bchart}[step=0.1,max=0.5]
        \bcbar[text=GPT-4o-mini, color=BlueViolet!20]{0.343693}
        \bcskip{0.1cm}
        \bcbar[text=GPT-4o, color=CornflowerBlue!20]{0.372943}
        \bcskip{0.1cm}
        \bcbar[text=GPT-4-turbo, color=Fuchsia!20]{0.33638}
        \bcskip{0.1cm}
        \bcbar[text=Gemini 2.0 flash, color=SpringGreen!20]{0.3766}
        \bcskip{0.1cm}
        \bcbar[text=instructblip-t5, color=RawSienna!20]{0.274223}
        \bcskip{0.1cm}
        \bcxlabel{Zero Shot (en)}
    \end{bchart}&  
        \begin{bchart}[step=0.1,max=0.5]
        \bcbar[text=GPT-4o-mini, color=BlueViolet!20]{0.276051}
        \bcskip{0.1cm}
        \bcbar[text=GPT-4o, color=CornflowerBlue!20]{0.310786}
        \bcskip{0.1cm}
        \bcbar[text=GPT-4-turbo, color=Fuchsia!20]{0.374771}
        \bcskip{0.1cm}
        \bcbar[text=Gemini 2.0 flash, color=SpringGreen!20]{0.372943}
        \bcskip{0.1cm}
        \bcbar[text=instructblip-t5, color=RawSienna!20]{}
        \bcskip{0.1cm}
            \bcxlabel{ICL (en)}
        \end{bchart} &
        \begin{bchart}[step=0.1,max=0.5]
        \bcbar[text=GPT-4o-mini, color=BlueViolet!20]{0.301645}
        \bcskip{0.1cm}
        \bcbar[text=GPT-4o, color=CornflowerBlue!20]{0.40585}
        \bcskip{0.1cm}
        \bcbar[text=GPT-4-turbo, color=Fuchsia!20]{0.334552}
        \bcskip{0.1cm}
        \bcbar[text=Gemini 2.0 flash, color=SpringGreen!20]{0.329068}
        \bcskip{0.1cm}
        \bcbar[text=instructblip-t5, color=RawSienna!20]{0.26691}
        \bcskip{0.1cm}
            \bcxlabel{First Describe (en)}
        \end{bchart}  \\
    \end{tabular}
}

\begin{center}
    \resizebox{10cm}{!}{
        \begin{tabular}{cc}
            \begin{bchart}[step=0.1,max=0.5]
            \bcbar[text=GPT-4o-mini, color=BlueViolet!20]{0.241316}
            \bcskip{0.1cm}
            \bcbar[text=GPT-4o, color=CornflowerBlue!20]{0.230347}
            \bcskip{0.1cm}
            \bcbar[text=GPT-4-turbo, color=Fuchsia!20]{0.197441}
            \bcskip{0.1cm}
            \bcbar[text=Gemini 2.0, color=SpringGreen!20]{0.151737}
            \bcskip{0.1cm}
            \bcbar[text=instructblip-t5, color=RawSienna!20]{0.274223}
            \bcskip{0.1cm}
                \bcxlabel{Wrong Image (en)}
            \end{bchart} &  
            \begin{bchart}[step=0.1,max=0.5]
            \bcbar[text=GPT-4o-mini, color=BlueViolet!20]{0.21755}
            \bcskip{0.1cm}
            \bcbar[text=GPT-4o, color=CornflowerBlue!20]{0.232176}
            \bcskip{0.1cm}
            \bcbar[text=GPT-4-turbo, color=Fuchsia!20]{0.294333}
            \bcskip{0.1cm}
            \bcbar[text=Gemini 2.0, color=SpringGreen!20]{0.159049}
            \bcskip{0.1cm}
            \bcbar[text=instructblip-t5, color=RawSienna!20]{}
            \bcskip{0.1cm}
                \bcxlabel{Without Image (en)}
            \end{bchart} 
        \end{tabular}
    }
\end{center}
    \caption{Performance comparison of each model across experiments on the \textbf{Art English} dataset}
    \label{fig:BarChart-Art-En-AllModelComparison}
\end{figure}
\begin{figure}[H]
\resizebox{\textwidth}{!}{
    \begin{tabular}{ccc}
    \begin{bchart}[step=0.1,max=0.5]
        \bcbar[text=GPT-4o-mini, color=BlueViolet!20]{0.310121}
        \bcskip{0.1cm}
        \bcbar[text=GPT-4o, color=CornflowerBlue!20]{0.413783}
        \bcskip{0.1cm}
        \bcbar[text=GPT-4-turbo, color=Fuchsia!20]{0.313293}
        \bcskip{0.1cm}
        \bcbar[text=Gemini 2.0 flash, color=SpringGreen!20]{0.435121}
        \bcskip{0.1cm}
        \bcxlabel{Zero Shot (fa)}
    \end{bchart}&  
        \begin{bchart}[step=0.1,max=0.5]
        \bcbar[text=GPT-4o-mini, color=BlueViolet!20]{0.224751}
        \bcskip{0.1cm}
        \bcbar[text=GPT-4o, color=CornflowerBlue!20]{0.385525}
        \bcskip{0.1cm}
        \bcbar[text=GPT-4-turbo, color=Fuchsia!20]{0.310121}
        \bcskip{0.1cm}
        \bcbar[text=Gemini 2.0 flash, color=SpringGreen!20]{0.377739}
        \bcskip{0.1cm}
            \bcxlabel{ICL (fa)}
        \end{bchart} &
        \begin{bchart}[step=0.1,max=0.5]
        \bcbar[text=GPT-4o-mini, color=BlueViolet!20]{0.312268}
        \bcskip{0.1cm}
        \bcbar[text=GPT-4o, color=CornflowerBlue!20]{0.422578}
        \bcskip{0.1cm}
        \bcbar[text=GPT-4-turbo, color=Fuchsia!20]{0.295822}
        \bcskip{0.1cm}
        \bcbar[text=Gemini 2.0 flash, color=SpringGreen!20]{0.504614}
        \bcskip{0.1cm}
            \bcxlabel{First Describe (fa)}
        \end{bchart}  \\
    \end{tabular}
}

\begin{center}
    \resizebox{10cm}{!}{
        \begin{tabular}{cc}
            \begin{bchart}[step=0.1,max=0.5]
            \bcbar[text=GPT-4o-mini, color=BlueViolet!20]{0.221886}
            \bcskip{0.1cm}
            \bcbar[text=GPT-4o, color=CornflowerBlue!20]{0.247721}
            \bcskip{0.1cm}
            \bcbar[text=$\quad$GPT-4-turbo, color=Fuchsia!20]{}
            \bcskip{0.1cm}
            \bcbar[text=Gemini 2.0, color=SpringGreen!20]{0.121107}
            \bcskip{0.1cm}
                \bcxlabel{Wrong Image (fa)}
            \end{bchart} &  
            \begin{bchart}[step=0.1,max=0.5]
            \bcbar[text=GPT-4o-mini, color=BlueViolet!20]{0.23515}
            \bcskip{0.1cm}
            \bcbar[text=GPT-4o, color=CornflowerBlue!20]{0.292162}
            \bcskip{0.1cm}
            \bcbar[text=GPT-4-turbo, color=Fuchsia!20]{0.213524}
            \bcskip{0.1cm}
            \bcbar[text=Gemini 2.0 flash, color=SpringGreen!20]{0.267589}
            \bcskip{0.1cm}
                \bcxlabel{Without Image (fa)}
            \end{bchart} 
        \end{tabular}
    }
\end{center}
    \caption{Performance comparison of each model across experiments on the \textbf{MEENA Farsi} dataset}
    \label{fig:BarChart-MEENA-Fa-AllModelComparison}
\end{figure}
\begin{figure}[H]
\resizebox{\textwidth}{!}{
    \begin{tabular}{ccc}
    \begin{bchart}[step=0.1,max=0.5]
        \bcbar[text=GPT-4o-mini, color=BlueViolet!20]{0.323583}
        \bcskip{0.1cm}
        \bcbar[text=GPT-4o, color=CornflowerBlue!20]{0.354662}
        \bcskip{0.1cm}
        \bcbar[text=GPT-4-turbo, color=Fuchsia!20]{0.305302}
        \bcskip{0.1cm}
        \bcbar[text=Gemini 2.0 flash, color=SpringGreen!20]{0.297989}
        \bcskip{0.1cm}
        \bcxlabel{Zero Shot (fa)}
    \end{bchart}&  
        \begin{bchart}[step=0.1,max=0.5]
        \bcbar[text=GPT-4o-mini, color=BlueViolet!20]{0.250457}
        \bcskip{0.1cm}
        \bcbar[text=GPT-4o, color=CornflowerBlue!20]{0.239488}
        \bcskip{0.1cm}
        \bcbar[text=GPT-4-turbo, color=Fuchsia!20]{0.305302}
        \bcskip{0.1cm}
        \bcbar[text=Gemini 2.0 flash, color=SpringGreen!20]{0.318099}
        \bcskip{0.1cm}
            \bcxlabel{ICL (fa)}
        \end{bchart} &
        \begin{bchart}[step=0.1,max=0.5]
        \bcbar[text=GPT-4o-mini, color=BlueViolet!20]{0.248629}
        \bcskip{0.1cm}
        \bcbar[text=GPT-4o, color=CornflowerBlue!20]{0.374771}
        \bcskip{0.1cm}
        \bcbar[text=GPT-4-turbo, color=Fuchsia!20]{0.265082}
        \bcskip{0.1cm}
        \bcbar[text=Gemini 2.0 flash, color=SpringGreen!20]{0.387569}
        \bcskip{0.1cm}
            \bcxlabel{First Describe (fa)}
        \end{bchart}  \\
    \end{tabular}
}

\begin{center}
    \resizebox{10cm}{!}{
        \begin{tabular}{cc}
            \begin{bchart}[step=0.1,max=0.5]
            \bcbar[text=GPT-4o-mini, color=BlueViolet!20]{0.193784}
            \bcskip{0.1cm}
            \bcbar[text=GPT-4o, color=CornflowerBlue!20]{0.171846}
            \bcskip{0.1cm}
            \bcbar[text=$\quad$GPT-4-turbo, color=Fuchsia!20]{0.184644}
            \bcskip{0.1cm}
            \bcbar[text=Gemini 2.0, color=SpringGreen!20]{0.122486}
            \bcskip{0.1cm}
                \bcxlabel{Wrong Image (fa)}
            \end{bchart} &  
            \begin{bchart}[step=0.1,max=0.5]
            \bcbar[text=GPT-4o-mini, color=BlueViolet!20]{0.206581}
            \bcskip{0.1cm}
            \bcbar[text=GPT-4o, color=CornflowerBlue!20]{0.182815}
            \bcskip{0.1cm}
            \bcbar[text=GPT-4-turbo, color=Fuchsia!20]{0.186472}
            \bcskip{0.1cm}
            \bcbar[text=Gemini 2.0 flash, color=SpringGreen!20]{0.244973}
            \bcskip{0.1cm}
                \bcxlabel{Without Image (fa)}
            \end{bchart} 
        \end{tabular}
    }
\end{center}
    \caption{Performance comparison of each model across experiments on the \textbf{Art Farsi} dataset}
    \label{fig:BarChart-Art-Fa-AllModelComparison}
\end{figure}

\begin{center}
\begin{table}[H]
\begin{tabular}{|lccccc|}
\hline
\rule{0pt}{2ex}
            Methods     & \multicolumn{1}{c}{\textbf{Mathematics}} & \multicolumn{1}{c}{\textbf{Natural Science}} & \multicolumn{1}{c}{\textbf{Social Science}} & \multicolumn{1}{c}{\textbf{Humanities}} & \multicolumn{1}{c|}{\textbf{Other}} \\
                 \hline
            
            \multicolumn{6}{l}{\rule{0pt}{2ex}\textbf{\textit{Zero-Shot}}}         \\
            \hline
            \rule{0pt}{2ex}
GPT-4o-mini      &                                 0.346418	& 0.619835 &	0.553571	& 0.417815	& 0.199475
                          \\
GPT-4o           &                                 0.460041 &	\textbf{0.675} &	\textbf{0.625} &	0.5397 &	0.263298
                           \\
GPT-4-Turbo      &                                0.39417 &	0.467532 &	0.363636 &	0.477707 &	0
                         \\
Gemini-2.0-flash &                                 \textbf{0.490677} &	0.661157 &	\textbf{0.625} &	\textbf{0.540827} &	\textbf{0.32021}
                           \\
instructblip-t5  &                                 0.179588	& 0.330579 &	0.303571 &	0.279958 &	0.178478
                           \\
               \hline
                   
                \multicolumn{6}{l}{\rule{0pt}{2ex}\textbf{\textit{First-Descibe}}}  \\
                \hline
                 \rule{0pt}{2ex}
                 
GPT-4o-mini      &                                 0.348381	& 0.528926 &	0.446429 &	0.408271 &	0.217848
                           \\
GPT-4o           &                                 0.436475 &	0.675 &	\textbf{0.678571} &	\textbf{0.535408} &	0.263298
                         \\
GPT-4-Turbo      &                                 0.339549 &	0.545455 &	0.517857 &	0.465536 &	0.215223
                           \\
Gemini-2.0-flash &                                 \textbf{0.478901} &	\textbf{0.570248} &	0.625 &	0.487805 &	\textbf{0.28084}
                           \\
instructblip-t5  &                                 0.155054	& 0.330579 &	0.25 &	0.26193 &	0.076115                           \\
\hline
                 \multicolumn{6}{l}{\rule{0pt}{2ex}\textbf{\textit{ICL}}}                                                                   \\
                 \hline
                 \rule{0pt}{2ex}
GPT-4o-mini      &                                 0.320903 &	0.454545 &	0.392857 &	0.33298	 & 0.181102
                        \\
GPT-4o           &                                 0.421001	& 0.528926	& \textbf{0.53571}4	& 0.40403	& \textbf{0.257218}
                           \\
GPT-4-Turbo      &                                 0.33366 &	0.528926 &	0.5 &	0.474019 &	0.23622 
                           \\
Gemini-2.0-flash &                                 \textbf{0.498528} &	\textbf{0.586777}	& 0.482143	& \textbf{0.50053} &	0.244094 \\
\hline
\multicolumn{6}{l}{\rule{0pt}{2ex}\textbf{\textit{Wrong Image}}} \\
\hline
\rule{0pt}{2ex}

GPT-4o-mini      &                                 0.270854 &	0.454545 &	0.464286 &	0.342524 &	0.034121
                          \\
GPT-4o           &                                 0.274779 &	0.404959 &	0.339286 &	0.340403 &	0.028871
                           \\
GPT-4-Turbo      &                                 0.245752	& 0.25	& 0.272727 &	0.364246	& 0
                           \\
Gemini-2.0-flash &                                 0.165849 &	0.289256 &	0.303571	& 0.236479	& 0.018373
                           \\
instructblip-t5  &                                 0.180569	& 0.264463 &	0.285714 &	0.265111 &	0.03937
                           \\
\hline
\multicolumn{6}{l}{\rule{0pt}{2ex}\textbf{\textit{Without Image}}} \\
\hline
\rule{0pt}{2ex}

GPT-4o-mini      &                                 0.276742 &	0.446281 &	0.392857 &	0.358431 &	0.020997
                           \\
GPT-4o           &                                 0.274779 &	0.404959 &	0.339286 &	0.340403 &	0.028871
                           \\
GPT-4-Turbo      &                                 0.26791 &	0.429752 &	0.446429 &	0.430541 &	0.031496
                           \\
Gemini-2.0-flash &                                 0.31894 &	0.330579 &	0.446429 &	0.415695 &	0.005249
                           \\
\hline           
\end{tabular}
\caption{Accuracy comparison of different models across different experiments and course subjects on the \textbf{English MEENA} dataset.}
\end{table}
\end{center}

\begin{center}
    \begin{table}[H]
\begin{tabular}{|lccccc|}
\hline
\rule{0pt}{2ex}
                Methods & \multicolumn{1}{c}{\textbf{Mathematics}} & \multicolumn{1}{c}{\textbf{Natural Science}} & \multicolumn{1}{c}{\textbf{Social Science}} & \multicolumn{1}{c}{\textbf{Humanities}} & \multicolumn{1}{c|}{\textbf{Other}} \\
                 \hline
                  \multicolumn{6}{l}{\rule{0pt}{2ex}\textbf{\textit{Zero-Shot}}}                                                                  \\
                  \hline
                 \rule{0pt}{3ex}
GPT-4o-mini      &                                 0.284702 &	0.439437 &	0.380435 &	0.341637 &	0.202091
                           \\
GPT-4o           &                                0.377134 &	0.55493	& 0.554348	& 0.462989	& \textbf{0.261324}
                           \\
GPT-4-Turbo      &                                 0.252174 &	0.422535 &	0.554348 &	0.379359	& 0.214286
                           \\
Gemini-2.0-flash &                                 \textbf{0.414493} &	\textbf{0.577465} &	\textbf{0.586957}	& \textbf{0.476868}	& 0.229965
                           \\
                           \hline

                 \multicolumn{6}{l}{\rule{0pt}{2ex}\textbf{\textit{First-Describe}}}                                                                     \\
                 \hline
                 \rule{0pt}{3ex} 
GPT-4o-mini      &                                 0.270946 &	0.401709 &	0	& 0.346692 &	0
                           \\
GPT-4o           &                                 0.385829	& 0.6	& \textbf{0.630435} &	0.469395 &	0.249129
                          \\
GPT-4-Turbo      &                                 0.238614 &	0.45283 &	0.484848 &	0.354093 &	0.203833
                           \\
Gemini-2.0-flash &                                 \textbf{0.509501} &	\textbf{0.566197} &	0.565217 &	\textbf{0.541281} &	\textbf{0.250871}
                           \\
                           \hline

                 \multicolumn{6}{l}{\rule{0pt}{2ex}\textbf{\textit{ICL}}}       
                 \\
                 \hline
                 \rule{0pt}{3ex}
GPT-4o-mini      &                                 0.242512	& 0.267606	& 0.271739 &	0.206762 &	0.183074
                          \\
GPT-4o           &                                 0.37037 &	0.490141 &	0.521739 &	0.411388 &	\textbf{0.254355}
                           \\
GPT-4-Turbo      &                                 0.249919 &	0.461972 &	0.51087 &	0.372598 &	0.203833
                           \\
Gemini-2.0-flash &                                 \textbf{0.509501} &	\textbf{0.566197} &	\textbf{0.565217} &	\textbf{0.541281} &	0.250871
                           \\
                           \hline

                 \multicolumn{6}{l}{\rule{0pt}{2ex}\textbf{\textit{Wrong Image}}}      
                 \\
                 \hline
                 \rule{0pt}{3ex}
GPT-4o-mini      &                                 0.202576 &	0.35493 &	0.25 &	0.260142 &	0.052265
                           \\
GPT-4o           &                                 0.22754 &	0.366667 &	0.378049 &	0.256077 &	0.115108
                           \\
Gemini-2.0-flash &                                 0.094042 &	0.267606 &	0.326087 &	0.148754 &	0.008711
                           \\
                           \hline
                 \multicolumn{6}{l}{\rule{0pt}{2ex}\textbf{\textit{Without Image}}}                                                                                                                                       \\
                 \hline
                 \rule{0pt}{3ex}
GPT-4o-mini      &                                 0.202254 &	0.397183	& 0.402174	& 0.286833 &	0.033101
                           \\
GPT-4o           &                                 0.226 &	0.442623 &	0 &	0.346743 &	0
                           \\
GPT-4-Turbo      &                                 0.150081 &	0.383099 &	0.456522 &	0.291815 &	0.029617
                           \\
Gemini-2.0-flash &                                 0.234461 &	0.388732 &	0.423913 &	0.330249	& 0.04007
                     \\
                     \hline
\end{tabular}
\caption{Accuracy comparison of different models across different experiments and course subjects on the \textbf{Farsi MEENA} dataset.}
\end{table}
\end{center}

\section{Prompts}
\label{app:prompts}
\begin{figure}[H]
    \begin{tcolorbox}[title=LLM as a Judge(Translation), colback=gray!5, colframe=black, fonttitle=\bfseries, coltitle=white, colbacktitle=black]
Welcome to the Translation Quality Assessment Tool. This tool is designed to evaluate the quality of translations from Persian to English. Please read each translated text carefully and use the rubric below to assign a score based on how well the translation preserves the meaning of the original text.\\

          Scoring Rubric:\\
          - 5 - Excellent: The translation conveys all aspects of the original meaning without any noticeable errors.\\
          - 4 - Good: Minor errors are present but do not alter the fundamental meaning.\\
          - 3 - Acceptable: Some parts of the meaning are lost or altered, but the overall intent is still recognizable.\\
          - 2 - Poor: Significant portions of the meaning are incorrect or missing.\\
          - 1 - Unacceptable: The translation fails to convey the original meaning.\\

          Please focus on the semantic accuracy of the translation. Minor grammatical or syntactic errors should be overlooked unless they significantly impact the overall understanding of the text.

\end{tcolorbox}
\end{figure}

\begin{figure}[H]
    \begin{tcolorbox}[title=LLM as a Judge(Answer Extraction), colback=gray!5, colframe=black, fonttitle=\bfseries, coltitle=white, colbacktitle=black]
Given a question and its answer, identify the selected option (1, 2, 3, or 4). 
If the choice is explicitly mentioned by number or implicitly indicated by the content, return the corresponding numerical option.
If the answer mentions it does not have the image, return "no\_image."
If the answer mentions it does not understand the image, return "cannot\_understand."
If the answer mentions that the image is wrong or not relevant to the question, return "wrong\_image."
If the selection cannot be determined or is ambiguous, return "unknown." 
Output only the number (1, 2, 3, or 4), "no\_image", "cannot\_understand", "wrong\_image", or "unknown" without any explanations.

\end{tcolorbox}
\end{figure}

\begin{figure}[H]
\begin{tcolorbox}[title= Persian to English translation prompts, colback=gray!5, colframe=black, fonttitle=\bfseries, coltitle=white, colbacktitle=black]
Please accurately translate the following multiple-choice question from Persian to English (Both Question and corresponding options). Use precise and professional terminology,
                  especially if specialized terms are involved. The question may be related to fields such as art, chemistry, physics, mathematics, biology, linguistics, geology, or other domains. Ensure that the translation is accurate and maintains the original meaning.\\
                  Question:\{mcq\_question\}\\

                  Please answer like this template:\\
                  Question:\\
                  Choice 1:\\
                  Choice 2:\\
                  Choice 3:\\
                  Choice 4:\\

\end{tcolorbox}   
\end{figure}

\begin{figure}[H]
\begin{tcolorbox}[title= English First Describe Prompt, colback=gray!5, colframe=black, fonttitle=\bfseries, coltitle=white, colbacktitle=black]
Below, you can see multiple-choice questions (with answers). \\
Question: \{mcq question\} \\
Choices: \\
1) \\
2) \\
3) \\
4) \\
Answer: Let's first describe the image carefully and provide all its details, then answer the question

\end{tcolorbox}   
\end{figure}

\begin{figure}[H]
\begin{tcolorbox}[title= English ICL Prompt, colback=gray!5, colframe=black, fonttitle=\bfseries, coltitle=white, colbacktitle=black]
Below, you can see multiple-choice questions (with answers). \\
Question: \{mcq question-shot1\} \\
Choices: \\
1) \\
2) \\
3) \\
4) \\
Answer: \{shot1 answer\}

Question: \{mcq question-shot2\} \\
Choices: \\
1) \\
2) \\
3) \\
4) \\
Answer: \{shot2 answer\}

Question: \{mcq question-shot3\} \\
Choices: \\
1) \\
2) \\
3) \\
4) \\
Answer: \{shot3 answer\}

Question: \{mcq question-shot4\} \\
Choices: \\
1) \\
2) \\
3) \\
4) \\
Answer: \{shot4 answer\}

Question: \{mcq question\} \\
Choices: \\
1) \\
2) \\
3) \\
4) \\
Answer: 

\end{tcolorbox}   
\end{figure}

\begin{figure}[H]
\begin{tcolorbox}[title= English Zero-Shot and Wronge-Image Prompt, colback=gray!5, colframe=black, fonttitle=\bfseries, coltitle=white, colbacktitle=black]
Below, you can see multiple-choice questions (with answers). \\
Question: \{mcq question\} \\
Choices: \\
1) \\
2) \\
3) \\
4) \\
Answer: 

\end{tcolorbox}   
\end{figure}

\begin{figure}[H]
\begin{tcolorbox}[title= English Without-Image Prompt, colback=gray!5, colframe=black, fonttitle=\bfseries, coltitle=white, colbacktitle=black]
Below, you can see multiple-choice questions (with answers). If no image is provided, choose the best choice based on the available information. \\
Question: \{mcq question\} \\
Choices: \\
1) \\
2) \\
3) \\
4) \\
Answer: 

\end{tcolorbox}   
\end{figure}


\section{Difficulty Level \& Trap Analysis}
\label{app:diff_lvl}

\begin{table}
\begin{tabular}{lllllll}
Model            & Experiment     & Easy    & Relatively Easy & Medium  & Relatively Difficult & Difficult \\
\textbf{\textit{Other}}            &                &         &                 &         &                      &           \\
gemini-2.0-flash & ICL            & 0.33333 & 0.22642         & 0.23762 & 0.26829              & 0.21341   \\
gemini-2.0-flash & first-describe & 0.32456 & 0.24528         & 0.21287 & 0.2439               & 0.25      \\
gemini-2.0-flash & zero-shot      & 0.32456 & 0.32075         & 0.22277 & 0.26829              & 0.13415   \\
gpt-4-turbo      & ICL            & 0.18421 & 0.35849         & 0.21287 & 0.2439               & 0.14634   \\
gpt-4-turbo      & first-describe & 0.17544 & 0.24528         & 0.19802 & 0.19512              & 0.21951   \\
gpt-4-turbo      & zero-shot      & 0.2193  & 0.26415         & 0.20297 & 0.19512              & 0.21341   \\
gpt-4o           & ICL            & 0.27193 & 0.28302         & 0.26238 & 0.2439               & 0.22561   \\
gpt-4o           & first-describe & 0.25439 & 0.26415         & 0.21287 & 0.2439               & 0.28659   \\
gpt-4o           & zero-shot      & 0.28947 & 0.28302         & 0.26238 & 0.19512              & 0.25      \\
gpt-4o-mini      & ICL            & 0.18421 & 0.11321         & 0.19802 & 0.17073              & 0.19512   \\
gpt-4o-mini      & first-describe & 0       & 0               & 0       & 0                    & 0         \\
gpt-4o-mini      & zero-shot      & 0.16667 & 0.13208         & 0.23267 & 0.19512              & 0.21341  \\
Human & & 0.58114 & 0.41528 & 0.44435 & 0.50975 & 0.53091
\end{tabular}

\caption{Comparision of different model performance by different experiment for "other" category}
\label{table:diff-other}
\end{table}

\begin{table}
\begin{tabular}{lllllll}
Model            & Experiment     & Easy    & Relatively Easy & Medium  & Relatively Difficult & Difficult \\
\textbf{\textit{Social Science}}   &                &         &                 &         &                      &           \\
gemini-2.0-flash & ICL            & 0.90909 & 0.66667         & 0.69799 & 0.56522              & 0.42063   \\
gemini-2.0-flash & first-describe & 0.72727 & 0.41667         & 0.65101 & 0.54348              & 0.46032   \\
gemini-2.0-flash & zero-shot      & 0.77273 & 0.41667         & 0.68456 & 0.54348              & 0.44444   \\
gpt-4-turbo      & ICL            & 0.72727 & 0.75            & 0.51007 & 0.34783              & 0.37302   \\
gpt-4-turbo      & first-describe & 0.75    & 0.5             & 0.53435 & 0.38462              & 0.32759   \\
gpt-4-turbo      & zero-shot      & 0.72727 & 0.58333         & 0.4698  & 0.3913               & 0.30952   \\
gpt-4o           & ICL            & 0.63636 & 0.66667         & 0.57047 & 0.47826              & 0.35714   \\
gpt-4o           & first-describe & 0.86364 & 0.58333         & 0.67114 & 0.52174              & 0.5       \\
gpt-4o           & zero-shot      & 0.72727 & 0.75            & 0.63087 & 0.54348              & 0.42063   \\
gpt-4o-mini      & ICL            & 0.45455 & 0.33333         & 0.32886 & 0.17391              & 0.19048   \\
gpt-4o-mini      & first-describe & 0.375   & 0.28571         & 0.48889 & 0.29167              & 0.39394   \\
gpt-4o-mini      & zero-shot      & 0.77273 & 0.5             & 0.48322 & 0.43478              & 0.3254  \\
Human & & 0.57772 & 0.48916 & 0.44785 & 0.33521 & 0.55833
\end{tabular}

\caption{Comparision of different model performance by different experiment for "Social Science" category}
\label{table:diff-social-science}
\end{table}

\begin{table}
\begin{tabular}{lllllll}
Model            & Experiment     & Easy    & Relatively Easy & Medium  & Relatively Difficult & Difficult \\
Natural Science &                &         &                 &         &                      &           \\
gemini-2.0-flash & ICL            & 0.61856 & 0.52747         & 0.42128 & 0.40678              & 0.33787   \\
gemini-2.0-flash & first-describe & 0.68041 & 0.58242         & 0.53386 & 0.49831              & 0.49134   \\
gemini-2.0-flash & zero-shot      & 0.66753 & 0.53297         & 0.46878 & 0.41695              & 0.40594   \\
gpt-4-turbo      & ICL            & 0.54897 & 0.46154         & 0.35972 & 0.35593              & 0.29208   \\
gpt-4-turbo      & first-describe & 0.54381 & 0.3956          & 0.34125 & 0.32881              & 0.28094   \\
gpt-4-turbo      & zero-shot      & 0.55155 & 0.42308         & 0.36412 & 0.37966              & 0.30817   \\
gpt-4o           & ICL            & 0.61856 & 0.45604         & 0.39314 & 0.42034              & 0.32426   \\
gpt-4o           & first-describe & 0.63918 & 0.58791         & 0.4635  & 0.46441              & 0.37129   \\
gpt-4o           & zero-shot      & 0.66753 & 0.53846         & 0.44943 & 0.44746              & 0.37252   \\
gpt-4o-mini      & ICL            & 0.35567 & 0.1978          & 0.19877 & 0.17627              & 0.15965   \\
gpt-4o-mini      & first-describe & 0.49141 & 0.41333         & 0.34447 & 0.35849              & 0.2535    \\
gpt-4o-mini      & zero-shot      & 0.50773 & 0.37363         & 0.32454 & 0.31864              & 0.28713  \\
Human & & 0.66489 & 0.50379 & 0.52244 & 0.43111 & 0.62153
\end{tabular}
\caption{Comparision of different model performance by different experiment for "Natural Science" category}
\label{table:diff-natural-science}
\end{table}

\begin{table}
\begin{tabular}{lllllll}
Model            & Experiment     & Easy    & Relatively Easy & Medium  & Relatively Difficult & Difficult \\
Mathematics      &                &         &                 &         &                      &           \\
gemini-2.0-flash & ICL            & 0.41679 & 0.33824         & 0.32695 & 0.23077              & 0.25979   \\
gemini-2.0-flash & first-describe & 0.53973 & 0.57353         & 0.52531 & 0.53846              & 0.45077   \\
gemini-2.0-flash & zero-shot      & 0.47676 & 0.48529         & 0.40903 & 0.50769              & 0.3618    \\
gpt-4-turbo      & ICL            & 0.27886 & 0.32353         & 0.25513 & 0.29231              & 0.20878   \\
gpt-4-turbo      & first-describe & 0.28395 & 0.30882         & 0.23966 & 0.22222              & 0.1966    \\
gpt-4-turbo      & zero-shot      & 0.29985 & 0.36765         & 0.23598 & 0.30769              & 0.22894   \\
gpt-4o           & ICL            & 0.41079 & 0.48529         & 0.36389 & 0.4                  & 0.33808   \\
gpt-4o           & first-describe & 0.43028 & 0.42647         & 0.37893 & 0.44615              & 0.35469   \\
gpt-4o           & zero-shot      & 0.43478 & 0.41176         & 0.37346 & 0.4                  & 0.33333   \\
gpt-4o-mini      & ICL            & 0.28036 & 0.27941         & 0.23598 & 0.27692              & 0.21827   \\
gpt-4o-mini      & first-describe & 0.31049 & 0.375           & 0.24368 & 0.34286              & 0.27126   \\
gpt-4o-mini      & zero-shot      & 0.30885 & 0.29412         & 0.27497 & 0.35385              & 0.27639   \\
Human & & 0.62299 & 0.45470 & 0.55884 & 0.37569 & 0.628374
\end{tabular}
\caption{Comparision of different model performance by different experiment for "Mathematics" category}
\label{table:diff-math}
\end{table}

\begin{table}[]
\begin{tabular}{lllllll}
Model            & Experiment     & Easy & Relatively Easy & Medium & Relatively Difficult & Difficult \\
Humanities       &                &      &                 &        &                      &           \\
gemini-2.0-flash & ICL            & 0.6  & 0.66667         & 0.8125 & 0.40625              & 0.36111   \\
gemini-2.0-flash & first-describe & 0.6  & 1               & 0.5625 & 0.6875               & 0.41667   \\
gemini-2.0-flash & zero-shot      & 0.6  & 0.66667         & 0.75   & 0.53125              & 0.55556   \\
gpt-4-turbo      & ICL            & 0.4  & 0.33333         & 0.625  & 0.5                  & 0.5       \\
gpt-4-turbo      & first-describe & 0    & 0.5             & 0.6    & 0.48148              & 0.45      \\
gpt-4-turbo      & zero-shot      & 0.4  & 0.66667         & 0.5625 & 0.625                & 0.5       \\
gpt-4o           & ICL            & 0.4  & 0.66667         & 0.5    & 0.59375              & 0.47222   \\
gpt-4o           & first-describe & 0.4  & 1               & 0.6875 & 0.71875              & 0.52778   \\
gpt-4o           & zero-shot      & 0.4  & 1               & 0.5625 & 0.59375              & 0.5       \\
gpt-4o-mini      & ICL            & 0.2  & 0.66667         & 0.4375 & 0.28125              & 0.16667   \\
gpt-4o-mini      & first-describe & 0    & 0               & 0      & 0                    & 0         \\
gpt-4o-mini      & zero-shot      & 0.2  & 0.33333         & 0.625  & 0.40625              & 0.27778  \\
Human & & 0.57200 & 0.56333 & 0.45625 & 0.52031 & 0.60138
\end{tabular}
\caption{Comparision of different model performance by different experiment for "Humanities" category}
\label{table:diff-math}
\end{table}

\begin{table}
\begin{tabular}{llll}
Model            & Experiment     & \% correct on Trap & \% correct on None-Trap \\
gemini-2.0-flash & first-describe & 0.4377             & 0.52437                 \\
gemini-2.0-flash & ICL            & 0.32068            & 0.39458                 \\
gemini-2.0-flash & zero-shot      & 0.3833             & 0.45042                 \\
gpt-4-turbo      & first-describe & 0.26381            & 0.30519                 \\
gpt-4-turbo      & ICL            & 0.27135            & 0.32157                 \\
gpt-4-turbo      & zero-shot      & 0.27641            & 0.32418                 \\
gpt-4o-mini      & first-describe & 0.26996            & 0.32428                 \\
gpt-4o-mini      & ICL            & 0.20304            & 0.23137                 \\
gpt-4o-mini      & zero-shot      & 0.27704            & 0.31989                 \\
gpt-4o           & first-describe & 0.37318            & 0.43716                 \\
gpt-4o           & ICL            & 0.3365             & 0.4                     \\
gpt-4o           & zero-shot      & 0.35863            & 0.43007  
                \\
human & & 0.53485 & 0.56999
\end{tabular}
\caption{Comparison of different models' performance on trap and non-trap questions.}
\label{table:performance-on-trap}
\end{table}

\end{document}